\documentclass[]{bytedance_seed}

\definecolor{color21}{RGB}{180, 230, 190}  
\definecolor{color22}{RGB}{240, 255, 240}  
\definecolor{color11}{RGB}{140, 190, 230}  
\definecolor{color12}{RGB}{230, 240, 255}  
\definecolor{color31}{RGB}{255, 190, 140}   
\definecolor{color32}{RGB}{255, 225, 210}  
\definecolor{color41}{RGB}{255, 245, 180}  
\definecolor{color42}{RGB}{255, 250, 210}  
\definecolor{color43}{RGB}{255, 255, 240}  

\definecolor{promptboxblue}{RGB}{59, 130, 246}
\definecolor{promptboxgray}{RGB}{107, 114, 128}
\definecolor{promptboxlightgray}{RGB}{243, 244, 246}
\definecolor{promptboxgreen}{RGB}{34, 197, 94}

\usepackage[toc,page,header]{appendix}

\usepackage{minitoc}
\usepackage{float}
\usepackage{amssymb}
\usepackage{pifont}
\usepackage{colortbl}
\usepackage{tcolorbox}
\usepackage{enumerate}
\usepackage{color} 
\usepackage{courier} 
\usepackage{wrapfig}
\usepackage{graphicx}
\usepackage{booktabs}   
\usepackage{graphicx}   
\usepackage{pifont}     
\usepackage{longtable}

\newcommand{\leanterm}[1]{{\fontfamily{pcr}\selectfont \colorbox{lightgray}{\textcolor{blue}{#1}}}}

\newcommand{\cmark}{\ding{51}}%
\newcommand{\xmark}{\ding{55}}%
\newtcolorbox{promptbox}[1][]{
  enhanced,
  breakable,
  colback=promptboxlightgray,
  colframe=promptboxblue!30,
  arc=8pt,
  boxrule=0.5pt,
  left=12pt,
  right=12pt,
  top=8pt,
  bottom=8pt,
  fonttitle=\bfseries,
  fontupper=\linespread{1.2}\selectfont,
  title=#1
}

\title{CriticLean: Critic-Guided Reinforcement Learning for Mathematical Formalization}

\affiliation[]{ByteDance Seed \quad \quad   Nanjing University}
\contribution{Full author list in Contributions}

\abstract{
Translating natural language mathematical statements into formal, executable code is a fundamental challenge in automated theorem proving. While prior work has focused on generation and compilation success, little attention has been paid to the \textit{critic phase}—the evaluation of whether generated formalizations truly capture the semantic intent of the original problem. In this paper, we introduce \textbf{CriticLean}, a novel critic-guided reinforcement learning framework that elevates the role of the critic from a passive validator to an active learning component. 
Specifically,
first, 
we propose the \textbf{CriticLeanGPT}, trained via supervised fine-tuning and reinforcement learning, to rigorously assess the semantic fidelity of Lean 4 formalizations. 
Then,
we introduce \textbf{CriticLeanBench}, a benchmark designed to measure models' ability to distinguish semantically correct from incorrect formalizations, and demonstrate that our trained CriticLeanGPT models can significantly outperform strong open- and closed-source baselines. Building on the CriticLean framework, we construct \textbf{FineLeanCorpus}, a dataset comprising over 285K problems that exhibits rich domain diversity, broad difficulty coverage, and high correctness based on human evaluation.
Overall, our findings highlight that optimizing the critic phase is essential for producing reliable formalizations and we hope our CriticLean will provide valuable insights for future advances in formal mathematical reasoning.

}

\date{July 8, 2025}
\correspondence{\email{liujiaheng@nju.edu.cn}, \email{zhangge.eli@bytedance.com}}

\checkdata[Project Page]{\url{https://github.com/multimodal-art-projection/CriticLean}}

\begin{document}\maketitle


\section{Introduction}
The formalization of mathematical statements \citep{yang2024formal} is a critical task in modern mathematical computation, particularly in the context of theorem provers like Lean 4 \citep{repl}. The translation of natural language mathematical problems into formal, executable code remains a significant challenge, as it requires not only syntactical accuracy but also a deep understanding of the problem's semantics \citep{scholze2022liquid, pfr_github}. Existing approaches have shown progress, but they often face limitations in accuracy, especially in the context of complex, high-level problems that involve sophisticated mathematical reasoning \citep{zheng2021minif2f, azerbayev2023proofnet, welleck2021, welleck2022, lewkowycz2022}.
 In contrast to existing works, we argue that the \textit{critic} phase—the step where the semantic correctness of generated formalizations is evaluated—is not only underexplored but also fundamentally essential to the success of mathematical autoformalization.

\begin{figure*}[t]
    \centering
    \includegraphics[width=0.95\linewidth]{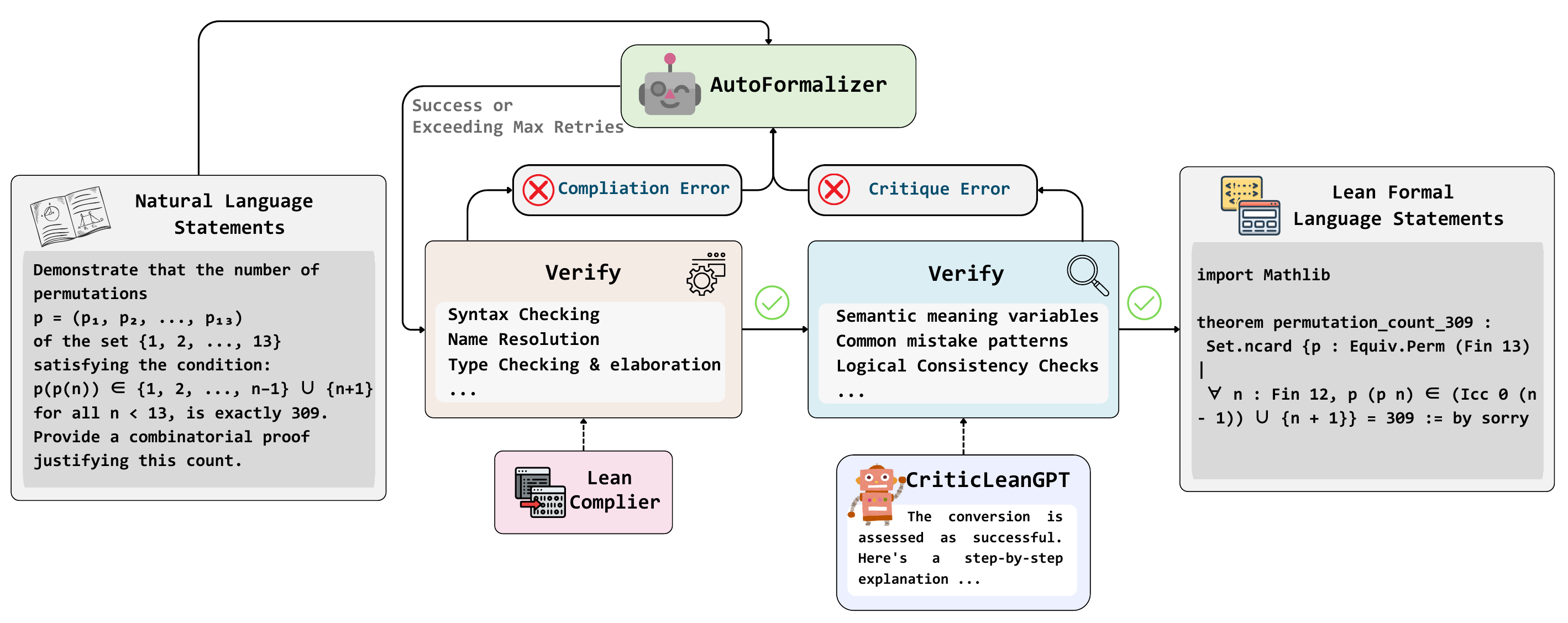}
    \caption{
    Illustration of CriticLean framework based on Critic-Guided Reinforcement Learning for Mathematics Autoformalization.} \label{fig:criticlean_intro}
\end{figure*}

Therefore,
in this paper, 
we systematically investigate and optimize the critic component and introduce \textbf{CriticLean}, a comprehensive framework that places the critic model at the center of the formalization pipeline. Unlike prior work that primarily focuses on generation quality or compiler validity, CriticLean introduces the reinforcement learning-based \textbf{CriticLeanGPT} models explicitly trained to evaluate whether the Lean 4 output truly reflects the intent of the original mathematical statement.,
and 
we present a full methodology for training, and evaluating CriticLeanGPT model.
Specifically,
as shown in Figure~\ref{fig:criticlean_intro},
for each natural language statement,
we apply the autoformalization iteratively based on the feedback from the Lean compiler and the CriticLeanGPT models, 
which is trained to critically assess whether a generated formalization accurately represents the semantics of the original mathematical statement. 
In each iteration, the feedback provides valuable signals that drive an iterative refinement process, further improving the quality of the final Lean code output.

Additionally, we present \textbf{CriticLeanBench}, a benchmark designed to evaluate the performance of CriticLeanGPT models,
which contain 500 natural language and Lean 4 language statement pairs (i.e., 250 correct and 250 incorrect pairs). Through extensive experiments, we demonstrate that our trained CriticLeanGPT models outperform the SOTA open-source models \citep{qwen, grattafiori2024llama, guo2025deepseek} and many closed-source API models \citep{gemini, openai2023gpt, guo2025seed1} greatly.

Furthermore, building upon our critic-centric CriticLean pipeline, we propose the high-quality open-source Lean 4 statement dataset \textbf{FineLeanCorpus},
comprising 285,957 fully verified entries. 
When compared to the previous related datasets (e.g., LeanWorkBook~\citep{ying2025leanworkbooklargescalelean}),
FineLeanCorpus is distinguished by its diversity in mathematical domains, difficulty distribution, and strict semantic validation via critical feedback loops. 
Notably, its difficulty distribution and targeted domain enrichment create a more structurally balanced training environment, mitigating overfitting and transforming sparse topics into well-supported sub-domains. Furthermore, to foster research into the upper echelons of mathematical reasoning, we have curated the specialized subset called FineLeanCorpus-Diamond, comprising over 36,000 high-difficulty problems.

.


\section{Related Works}
\subsection{Autoformalization}
Autoformalization~\cite{szegedy2020promising, wu2022autoformalization, yu2025generating} refers to the process by which AI systems parse natural language (NL) contents and translate them into machine-verifiable formal representations, such as those in theorem provers like Lean4~\citep{10.1007/978-3-030-79876-5_37} or Isabelle~\citep{10.5555/1791547}. 
Recent advances leverage large language models (LLMs)~\cite{zhang2024map} to tackle this problem through two primary paradigms: (1) In-context learning~\cite{wei2022chain}, where models utilize annotated examples~\cite{wu2022autoformalization, liu2023fimo, lu2024processdrivenautoformalizationlean4} to generate formalizations without explicit fine-tuning, (2) Supervised fine-tuning (e.g., ~\cite{lin2025goedel, xin2024deepseek, yu2025formalmath}), which adapts general-purpose LLMs into domain-specific autoformalization experts.
To assess correctness, prior works~\cite{lin2025goedel, xin2024deepseek, yu2025formalmath} employ LLM-as-judge~\cite{zheng2023judging} to verify semantic alignment between formal and informal statements.
We advance this by training the first open-sourced, domain-specific light LLM on top of Qwen~\cite{qwen3technicalreport} family for critiquing Lean4 statement alignment via reinforcement learning~\cite{shao2024deepseekmath}, 
enhancing both critique capability and formalization robustness.


\subsection{RL for LLM Reasoning}
The exploration of complex reasoning capabilities in Large Language Models (LLMs) has achieved significant advancements~\citep{liu2024ddk,liu2025comprehensive,Wang2025ReinforcementLO,Huang2025ThinkJLT,liu2024m2rc,he2025largelanguagemodelsdetect}, with Reinforcement Learning (RL) establishing itself as a critical paradigm for transcending the constraints inherent to Supervised Fine-Tuning (SFT)~\citep{yu2025dapo, yue2025vapo, wang2025reinforcement,liu2025prorl, liu2024m2rc}. Notable methodologies, including GRPO~\citep{shao2024deepseekmath,guo2025deepseek}, DAPO~\citep{yu2025dapo} have demonstrated substantial gains in mathematical reasoning and intricate problem-solving domains. However, the underlying mechanisms by which RL enhances reasoning capabilities remain insufficiently characterized. Empirical analyses increasingly indicate that RL primarily functions to activate, refine, or optimize the sampling of latent reasoning competencies rather than generating entirely novel cognitive frameworks de novo. For example, \citet{yue2025does} interrogate whether current reinforcement learning frameworks incorporating verifiable rewards (RLVR) genuinely expand the frontier of reasoning performance or merely enhance the efficiency of retrieving pre-existing solutions. 

\section{CriticLeanGPT}
\subsection{CriticLeanBench}

\subsubsection{Overview of CriticLeanBench}
\begin{figure*}[t]
    \centering
    \includegraphics[width=1.0\linewidth]{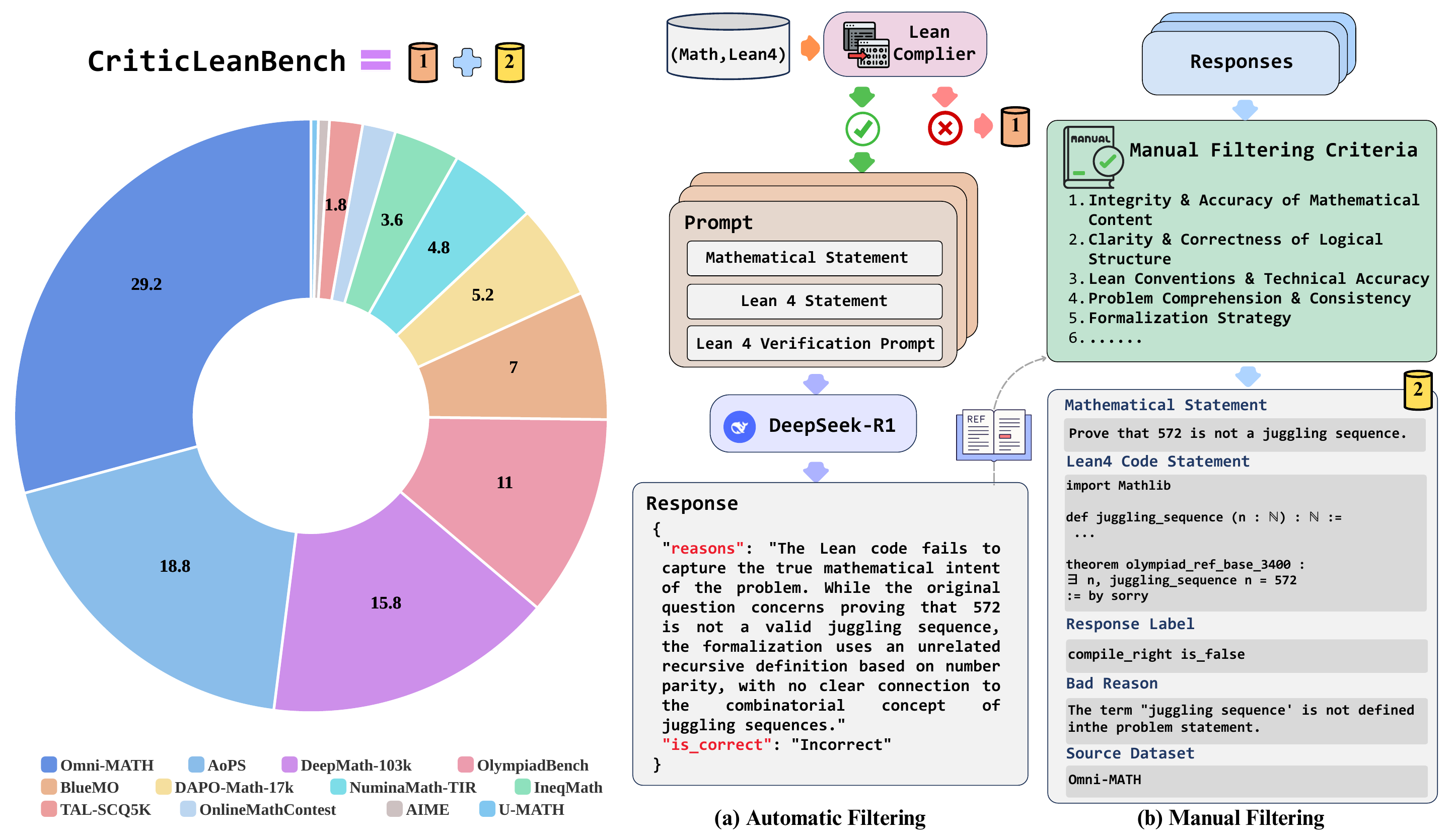}
    \caption{An overview for the CriticLeanBench construction.} \label{fig:criticleanbench}
\end{figure*}

CriticLeanBench aims to evaluate the critical reasoning of LLMs in key aspects such as translating natural language mathematical statements into formally verified theorem declarations in Lean 4, including critique and correction. By integrating these core dimensions, CriticLeanBench can comprehensively measure the performance of models in Formalization tasks. In this section, we will elaborate on the construction principles and processes of CriticLeanBench; the construction process is shown in Figure \ref{fig:criticleanbench}.

CriticLeanBench is constructed following the following principles: (1) It covers various types of errors, aiming to thoroughly evaluate the comprehensive capabilities of LLMs in capturing the semantic intent of formalized statements, using Template \ref{appendix:Formal Verification} for systematic assessment; (2) It incorporates diverse data sources to enhance the diversity and representativeness of evaluation data; and (3) It ensures the reliability and validity of the evaluation benchmark through a combination of expert review and automated verification.
\subsubsection{Automatically verified}
This section outlines the methodology employed to apply the Automatic Validation Filter , guided by predefined criteria for compiler and model-based verification.

\textbf{Data Collection} We selected Math Statements from data sources~\citep{
perez2025ai4math, mahdavi2025leveraging, yu2025dapo, he2025deepmath, numina_math_datasets, OnlineMathContest, HME100K_dataset}, such as Omni-MATH \citep{gao2024omni}, AIME \citep{zheng2021minif2f}, U-MATH \citep{chernyshev2024u}, DEMI-MathAnalysis \citep{demidovich1964problems}, HARDMath \citep{fan2410hardmath}, OlympiadBench \citep{he2024olympiadbench}, and BlueMO~\citep{bluemo2024}, along with their corresponding Lean 4 Statements from public dataset \citep{yu2025formalmath}.

\textbf{Compiler Verified} 
We submitted the Lean 4 statements from the above dataset to the Lean 4 compiler. If the compilation succeeded, the results were passed to the DeepSeek R1 \citep{guo2025deepseek} model for further processing.

\begin{itemize}
    \item \textbf{Compile false:} For statements that failed to compile, we randomly sampled 50 entries and retained the compiler feedback messages, which were included as part of our CriticLeanBench benchmark.
\end{itemize}

\textbf{LLM Verified} 
For the data that has passed compilation, we utilize a template \ref{appendix:Formal Verification} to process each sample's Math Statement and its corresponding Lean 4 Code Statement through the DeepSeek R1 large language model. The model is tasked with determining whether the Lean 4 Code Statement is consistent with the Math Statement, producing for each sample a tag indicating consistency or inconsistency along with a reasoning statement. To extract structured outputs from the model's responses, we employ regular expression-based pattern matching. The results are then submitted to human reviewers for further validation.

\subsubsection{Human Validation Filter}
This section outlines the methodology employed to apply the Human Validation Filter, guided by predefined criteria(detailed in Appendix~\ref{appendix:Formalization_Quality_Assessment_Criteria}). 

We categorize the data into two groups based on whether the \textbf{Lean-compiled} autoformalization output semantically aligns with the refined statement. Therefore, a Lean-compiled autoformalization falls into the following:
\begin{itemize}
    \item \textbf{Human Check right:} If the autoformalized statement \emph{both} compiles \emph{and} accurately captures the mathematical meaning, logic, and intended behavior of the original problem, it satisfies the semantic consistency criteria.
    \item \textbf{Human Check false:} If the autoformalized statement compiles but \emph{fails} to accurately represent the original mathematical problem's semantics. It violates one or more of the semantic consistency criteria, despite being syntactically valid Lean code. This often happens when there's a mismatch between the code's logic and the intended mathematical meaning.
\end{itemize}
Within the two categories, we select representative samples that mirror the characteristics of the original autoformalization statements while maximizing diversity within each subset and capturing the full spectrum of observed error types in the negative samples. This is achieved by balancing several factors:

\textbf{Stratified Representation:} The selected samples should maintain a similar distribution across different strata of the original dataset. These strata are defined by:
\begin{itemize}
\item {Problem Complexity:} This encompasses factors like the number of variables, quantifiers, logical connectives, and the depth of nested mathematical structures. 
\item {Mathematical Branches:} The samples should represent the various mathematical domains present in the original data.
\item {Statement Well-Formedness:} This refers to the degree to which the original mathematical statements adhere to standard mathematical notation and conventions. This stratification ensures that the subsets reflect the variability in the quality of the original problem statements.
\end{itemize}

\begin{figure*}[h]
    \centering
    \includegraphics[width=0.9\textwidth]{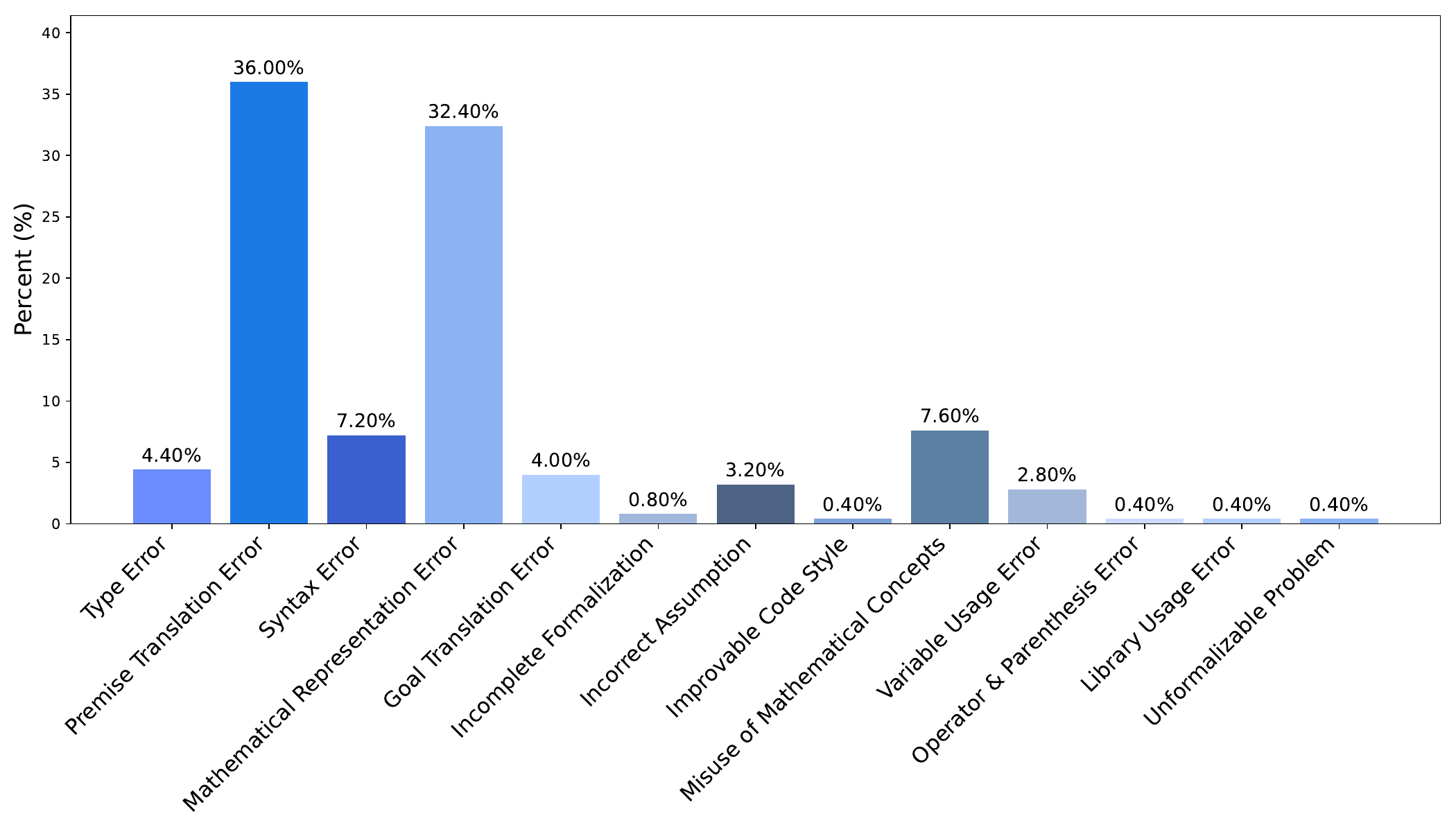}
    \caption{Distribution of Different Error Types of CriticLeanBench.} %
    \label{fig:Distribution of Different Error Types of CriticLeanBench}
\end{figure*}

\textbf{Comprehensive Error Coverage:} The negative samples are specifically chosen to exemplify the full range of typical errors observed in the autoformalization process. This range includes fundamental semantic and logical issues, such as Premise Translation Errors (e.g., incorrect domains or missing conditions), Mathematical Representation Errors (e.g., faulty expressions or definitions), and Goal Translation Errors. The set also covers issues such as Incorrect Assumptions, logical flaws like Operator \& Parenthesis Errors (e.g., misplaced quantifiers), and high-level structural problems like Incomplete Formalization, where crucial context is omitted(detailed in figure\ref{fig:Distribution of Different Error Types of CriticLeanBench}).

\subsubsection{Data Statistics}
\begin{table}[htbp]
\centering
\begin{minipage}{0.48\textwidth}
\centering
\begin{tabular}{cc}
\toprule
\textbf{Statistics} & \textbf{Number} \\
\midrule
\textbf{\#Problems} & $500$ \\
\textbf{Correct Pairs} & $250$ \\
~~~~- \textit{human check}   & $250$\\
\textbf{Incorrect Pairs} & $250$ \\
~~~~- \textit{human/compiler check}   & $200$/$50$\\
\midrule
\textbf{Question Tokens Length} \\
~~~~- \textit{max/min/avg}   & $1,583$/$495$/$700.94$ \\
\bottomrule
\end{tabular}
\end{minipage}
\hfill
\begin{minipage}{0.48\textwidth}
\centering
\begin{tabular}{ccccc}
\toprule
\textbf{Benchmark}  & \textbf{Critic} & \textbf{Lean} & \textbf{Test} \\
\midrule
CriticBench & \cmark & \xmark & \cmark \\
CodeCriticBench & \cmark & \xmark & \cmark \\
LLaVA-Critic & \cmark & \xmark & \xmark \\
\textbf{CriticLeanBench (ours)} & \cmark & \cmark & \cmark \\
\bottomrule
\end{tabular}
\end{minipage}
\caption{Dataset statistics and comparison of various code benchmark datasets.}
\label{tab:combined}
\end{table}

Table \ref{tab:combined} (left) presents the dataset statistics for CriticLeanBench. The dataset contains a total of 500 problems, comprising 250 correct and 250 incorrect Q/GT (Question and GT label) pairs distributed across the collection. The questions exhibit a notable variation in length, with a maximum token count of 1,583 and a minimum of 495 tokens, calculated using the Qwen2.5 \citep{qwen} tokenizer. On average, each question contains approximately 700.94 tokens. This characteristic distinguishes CriticLeanBench as a complex evaluation benchmark that requires models to process relatively lengthy inputs compared to those in standard datasets.

\subsubsection{Comparison to Other Benchmarks}

In Table \ref{tab:combined} (right), CriticLeanBench has the following features:
(1) We focus on a modest data size of 500 samples, acknowledging the expensive manual annotation costs (ranging from tens to hundreds of dollars per instance) and prioritizing efficiency without compromising evaluation rigor; (2) We integrate both Critic and Lean functionalities, distinguishing ourselves from benchmarks like CriticBench \citep{lin2024criticbench} and CodeCriticBench \citep{zhang2025codecriticbench} that lack Lean capabilities, and LLaVA-Critic \citep{xiong2025llava} that omits both Lean capabilities and test evaluation components; (3) We incorporate a test component focused on translating mathematical statements into formally verified theorem declarations in Lean 4, offering a fine-grained evaluation framework to assess the correctness of model transformations. This aspect of evaluative completeness remains unmatched by existing datasets.

\section{CriticLeanInstruct}
\label{sec:criticleaninstruct}
To enhance the CriticLeanGPT model's efficacy in critically evaluating the successful transformation of mathematical statements into Lean code, we constructs a comprehensive training dataset comprising 48,000 samples. This dataset consists of three high-quality components, designed to comprehensively improve the model's critical thinking and reasoning capabilities.

To significantly broaden the CriticLeanGPT model's knowledge coverage, we also integrates three times additional code and math \citep{openr1} datasets, enabling it to more comprehensively understand mathematical concepts and Lean code structures. 

Specifically, the Seed Data, with a 1:3 mix of math and code, is termed \textbf{CriticLeanInstruct(16K)}. When combined with data augmentation while retaining the same ratio, it forms the full \textbf{CriticLeanInstruct} dataset.

\subsection{Seed Data}
\label{sec:seed_data}
The seed data comprises 4,000 samples evenly split into 2,000 correct and 2,000 incorrect instances. For both correct and incorrect samples, human experts provided critical feedback \ref{appendix:Critical_Feedback_to_CoT}. Additionally, the incorrect samples include compiler error messages generated by the Lean 4 Compiler as supplementary feedback.
Then, we adopt the Gemini2.5-Pro~\citep{gemini} to extend the critical feedback to detailed Chain-of-Thought (CoT) explanations.

\begin{figure*}[h]
    \centering
    \includegraphics[width=0.9\textwidth]{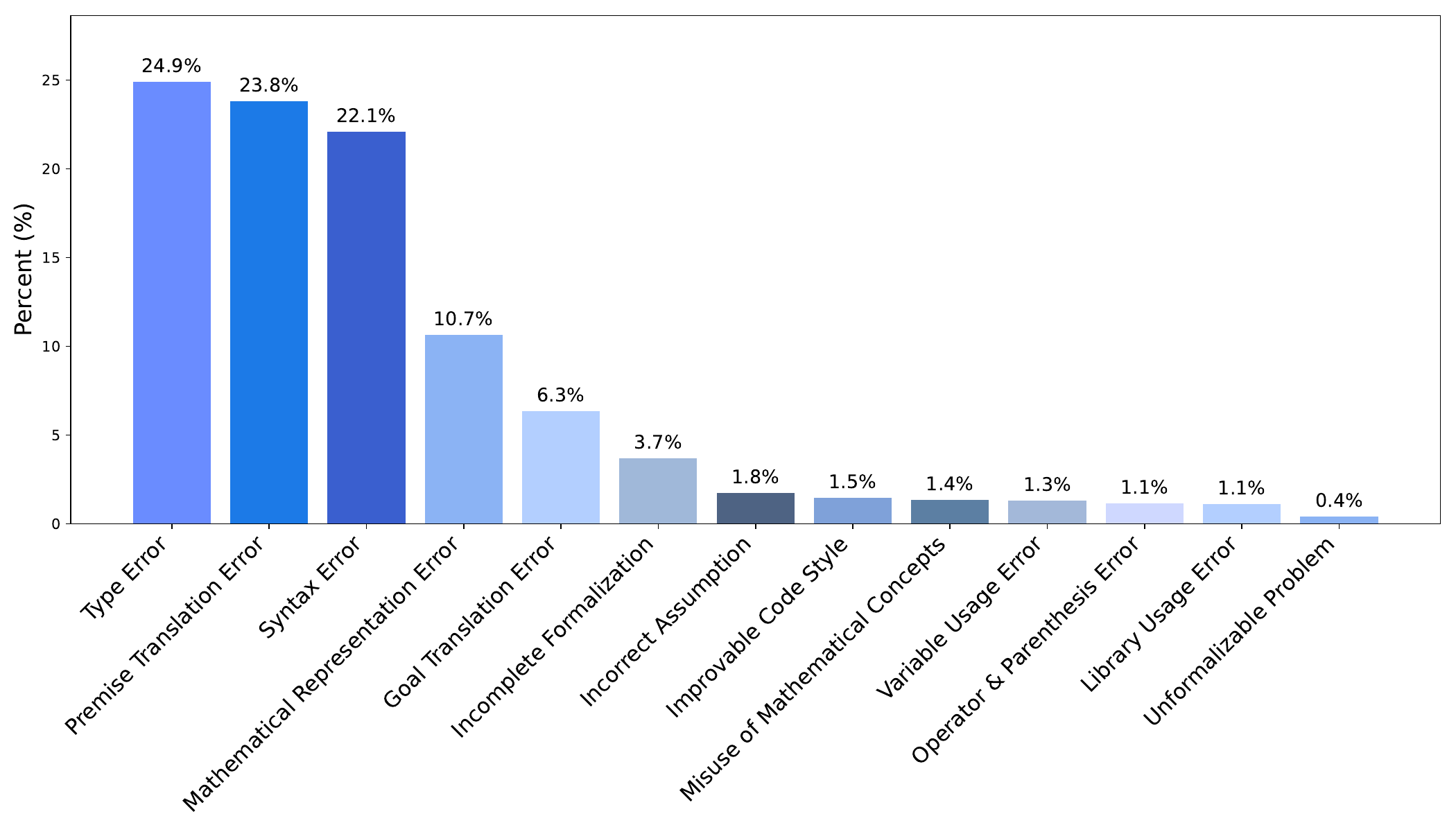}
    \caption{Distribution of Different Error Types.} %
    \label{fig:error_distribution}
\end{figure*}
To understand the primary failure modes, we analyzed the error distribution across the 2,000 incorrect samples in our seed data as illustrated in Figure~\ref{fig:error_distribution}. The analysis reveals two primary challenges:(1) Syntactic Barriers: The most frequent obstacles are syntactic, such as Type Error and  Syntax Errors. These issues typically prevent the code from compiling, indicating a fundamental difficulty in mastering the formal language.(2) Semantic Gaps: Beyond syntax, these errors stem from a failure to interpret the natural language scenario and accurately model its key information. This is evident in high rates of errors when translating a problem's core logical components—including its premises, goal, and mathematical representation. This semantic challenge is particularly acute in application-style "word problems," where difficulty comprehending complex contexts leads to a fundamentally flawed formalization.

Drawing from the error taxonomy (detailed in Appendix~\ref{appendix:Error Taxonomy}) and building upon our initial annotation standards (detailed in Appendix~\ref{appendix:Formalization_Quality_Assessment_Criteria}), we established a fine-grained checklist, which is provided in full in Appendix~\ref{appendix:checklist}. This checklist formalizes the observed error paradigms, providing the foundational framework for methodically constructing the negative samples used throughout our study.
\subsection{Data Augmentation}

\subsubsection{Correct Samples}
We selected \textbf{5560} correct mathematical statements and Lean code pairs from the FormalMATH~\citep{yu2025formalmath} dataset and used the Gemini-2.5-Pro model to generate Critical Chain-of-Thought for initial assessment. To ensure data quality, we kept these samples where the model's judgment is ``correct.''
\subsubsection{Incorrect Samples}
We provide two strategies to obtain the incorrect samples as follows:

\begin{itemize}
    \item Based on OmniMath, we adopt the Kimina-Autoformalizer-7B~\citep{kimina_prover_2025} to generate the Lean 4 code statements, and  we kept 2,000 Lean code snippets that failed to compile due to syntactic or logical issues during automated formalization processes. For each detected error, a detailed Chain-of-Thought (CoT) explanation was generated to elucidate the error's cause, enabling the model to recognize common compilation error patterns and thereby enhance its understanding of Lean 4 code syntax.
    \item Based on correct mathematical statements and Lean 4 code from the FormalMATH~\citep{yu2025formalmath} dataset, we implemented a three-step collaborative process to generate negative samples, aiming to enhance the CriticLeanGPT model's ability to identify subtle errors and logical flaws. First, a checklist of various potential issues, refined by human experts, was established as shown in Appendix~\ref{appendix:checklist}. Second, the Gemini 2.5 Pro model was invoked to randomly select error types from this checklist and modify correct Lean code accordingly, generating incorrect samples(detailed in Appendix~\ref{appendix:Lean Flaw Injection}). 
    Then, we adopt the Gemini model to generate the critical Chain-of-Thought explaination.
    %
\end{itemize}






\subsection{Training Paradigm}

\paragraph{Supervised fine-tuning (SFT).} These models are instruction-tuned versions based on their respective pre-trained Qwen2.5 checkpoints, with a focus on improving their ability to interpret and formalize complex mathematical statements expressed in natural language. The CriticLeanInstruct \ref{sec:criticleaninstruct} dataset includes Critic data consisting of mathematical statements converted into formally verified theorem declarations in Lean 4, along with three times as much code and mathematics data for SFT (Supervised Fine-Tuning). We used the LLaMA-Factory \citep{zheng2024llamafactory} framework to facilitate the fine-tuning process and optimize model performance.

\paragraph{Reinforcement Learning Optimization (RL).} The recent success of R1-style methods has demonstrated the effectiveness of online RL using discrete, rule-based rewards \cite{deepseek-math}. In our pipeline, Qwen2.5 series and Qwen3 \citep{qwen3technicalreport} series are further refined using reinforcement learning signals derived from both format validation of the generated critic data and consistency checking between model predictions and expert-labeled ground truth (GT) labels . Specifically, the RL training data consists of 4,000 Seed Data \ref{sec:seed_data}, where each example transforms a mathematical statement into a corresponding formal proof in Lean 4. Based on this dataset, we apply a rule-based RL approach to optimize the model's capability in judgment reasoning. More specifically, we mainly utilize the GRPO \citep{shao2024deepseekmath} algorithm within the VeRL \citep{sheng2024hybridflow} reinforcement learning framework, whose optimization objective is:

\vspace{-5mm}
\noindent
\begin{small}
\begin{align}
& J_{\text{online}}(\pi_\theta; \mathcal{D}) = \nonumber \\
&\:\: \mathbb{E}_{x \sim \mathcal{D},\{y_i\}_{i=1}^{G} \sim \pi_{\theta_{\text{old}}} (y|x)} \left[ \frac{1}{G} \sum_{i=1}^{G} \min \left( \frac{\pi_{\theta}(y_i|x)}{\pi_{\theta_{\text{old}}} (y_i|x)} A_i, \right. \right. \nonumber \\
&\:\: \left. \left. \text{clip} \left( \frac{\pi_{\theta}(y_i|x)}{\pi_{\theta_{\text{old}}} (y_i|x)}, 1 - \epsilon, 1 + \epsilon \right) A_i \right) - \beta D_{\text{KL}} (\pi_{\theta} || \pi_{\text{ref}}) \right]
\end{align}
\end{small}
\vspace{-4mm}

\noindent
where $G$ is group size, and $A_i$ is advantage. The reward function is designed as follows\footnote{We do not include a length penalty in rewards to encourage longer thinking.}:

\vspace{-4mm}
\begin{small}
\begin{align}
r_{\text{accuracy}} &= \begin{cases}
    1, & \text{if } \text{judgement} = \text{label}\\
    0, & \text{if } \text{judgement} \neq \text{label}
\end{cases} \\
r_{\text{format}} &= \begin{cases}
    1, & \text{if } \text{format is right}\\
    0, & \text{if } \text{format is wrong}
\end{cases} \\
r_{\text{final}} &= \operatorname{min}(r_{\text{accuracy}}, r_{\text{format}})
\end{align}
\end{small}
\vspace{-4mm}
\section{Experiments}

\subsection{Experimental Setup} 
\textbf{Baseline Models.} For the closed-sourced API models, we select the following models: Claude35\_Sonnet2 \citep{claude35addendum}, Doubao-1.5-pro-32k \citep{guo2025seed1}, Gemini 2.5 Pro \citep{team2023gemini}, GPT-4o-2024-11-20 \citep{openai2023gpt}. We select a group of the most advanced open-source LLMs to serve as critic models for evaluation, which includes various reasoning models (DeepSeek-R1 \citep{guo2025deepseek}, QwQ-32B \citep{qwq32b}, Qwen3-8B \citep{qwen3technicalreport}, Qwen3-14B \citep{qwen3technicalreport}, Qwen3-32B \citep{qwen3technicalreport}), DeepSeek-Prover models \citep{xin2024deepseekproverv15harnessingproofassistant} (DeepSeek-Prover-V1.5-RL, DeepSeek-Prover-V1.5-SFT), Llama-3.3-70B-Instruct \citep{grattafiori2024llama} and several Qwen models (Qwen2.5-Coder-7B-Instruct \citep{hui2024qwen2}, Qwen2.5-Coder-32B-Instruct \citep{hui2024qwen2}, Qwen2.5-7B-Instruct \citep{qwen2.5}, Qwen2.5-14B-Instruct \citep{qwen2.5} and Qwen2.5-32B-Instruct \citep{qwen2.5}). 

\textbf{CriticLeanGPT Models.} We further evaluate three variants from the Qwen2.5 series and Qwen3 series. These are instruction-based models fine-tuned specifically on either the CriticLeanInstruct \ref{sec:criticleaninstruct} dataset or the RL-based clean critic Seed Data \ref{sec:seed_data}. All open-source models are inferred using the vLLM \citep{kwon2023efficient} framework with default inference parameters.

\subsection{Main Results}

\subsubsection{Evaluation of Critic Capability}
\begin{table}[!htp]
\centering
\begin{tabular}{cccccccc}
\toprule
Model & ACC & TPR & FPR & TNR & FNR & Reasoning & Open-source \\
\midrule
\multicolumn{8}{c}{\textbf{\textit{SOTA LLMs}}} \\
\midrule
\rowcolor{color12}
Gemini-2.5-Pro & \boxed{89.2}& \underline{95.6}& \underline{4.4}& \underline{82.8}& \underline{17.2}& \cmark& \xmark \\
\rowcolor{color12}
QwQ-32B & \underline{86.4}& 93.6& 6.4& 79.2& 20.8& \cmark& \cmark \\
\rowcolor{color12}
Qwen3-32B & 85.6& \textbf{96.0}& \textbf{4.0}& 75.2& 24.8& \cmark& \cmark \\
\rowcolor{color12}
Qwen3-235B-A22B & 84.8& 90.4& 9.6& 79.2& 20.8& \cmark& \cmark \\
\rowcolor{color12}
DeepSeek-R1 & 84.0& 90.8& 9.2& 77.2& 22.8& \cmark& \cmark \\
\rowcolor{color12}
Qwen3-14B & 83.6& 92.4& 7.6& 74.8& 25.2& \cmark& \cmark \\
\rowcolor{color12}
Qwen3-8B & 79.8& 94.4& 5.6& 65.2& 34.8& \cmark& \cmark \\
\rowcolor{color12}
Doubao-1.5-pro-32k & 78.4& 95.2& 4.8& 61.6& 38.4& \xmark& \xmark \\
\rowcolor{color12}
Claude35-Sonnet & 74.2& \boxed{97.2}& \boxed{2.8}& 51.2& 48.8& \xmark& \xmark \\
\rowcolor{color12}
Qwen2.5-32B-Instruct & 73.0& 91.6& 8.4& 54.4& 45.6& \xmark& \cmark \\
\rowcolor{color12}
Qwen2.5-Coder-32B-Instruct & 71.6& 91.6& 8.4& 51.6& 48.4& \xmark& \cmark \\
\rowcolor{color12}
Llama-3.3-70B-Instruct & 68.2& 95.2& 4.8& 41.2& 58.8& \xmark& \cmark \\
\rowcolor{color12}
GPT-4o-2024-11-20 & 67.8& 95.6& 4.4& 40.0& 60.0& \xmark& \xmark \\
\rowcolor{color12}
Qwen2.5-14B-Instruct & 66.6& 80.4& 19.6& 52.8& 47.2& \xmark& \cmark \\
\rowcolor{color12}
Qwen2.5-Coder-7B-Instruct & 65.4& 88.4& 11.6& 42.4& 57.6& \xmark& \cmark \\
\rowcolor{color12}
Qwen2.5-7B-Instruct & 60.8& 89.6& 10.4& 32.0& 68.0& \xmark& \cmark \\
\rowcolor{color12}
DeepSeek-Prover-V1.5-SFT & 52.4& 78.8& 21.2& 26.0& 74.0& \xmark& \cmark \\
\rowcolor{color12}
DeepSeek-Prover-V1.5-RL & 50.0& 76.4& 23.6& 23.6& 76.4& \xmark& \cmark \\

\midrule
\multicolumn{8}{c}{\textbf{\textit{CriticLeanGPT \textbf{(Ours)}}}} \\
\midrule
\rowcolor{color22}
Qwen3-8B-RL & 79.8& 90.0& 10.0& 72.0& 28.0& \cmark& \cmark \\
\rowcolor{color22}
Qwen3-14B-RL & 84.8& 91.6& 8.4& 78.0& 22.0& \cmark& \cmark \\
\rowcolor{color22}
Qwen3-32B-RL & \textbf{87.0}& 88.4& 11.6& \boxed{85.6}& \boxed{14.4}& \cmark& \cmark \\
\rowcolor{color22}
Qwen2.5-7B-Instruct-RL & 68.6& 85.6& 14.4& 51.6& 48.4& \xmark& \cmark \\
\rowcolor{color22}
Qwen2.5-14B-Instruct-RL & 69.4& 85.2& 14.8& 53.6& 46.4& \xmark& \cmark \\
\rowcolor{color22}
Qwen2.5-32B-Instruct-RL & 72.0& 60.4& 39.6& \textbf{83.6}& \textbf{16.4}& \xmark& \cmark \\
\rowcolor{color31}
Qwen2.5-7B-Instruct-SFT & 69.8& 94.4& 5.6& 45.2& 54.8& \xmark& \cmark \\
\rowcolor{color31}
Qwen2.5-14B-Instruct-SFT & 70.6& 83.6& 16.4& 57.6& 42.4& \xmark& \cmark \\
\rowcolor{color31}
Qwen2.5-32B-Instruct-SFT & 76.2& 85.2& 14.8& 67.2& 32.8& \xmark& \cmark \\
\rowcolor{color32}
Qwen2.5-7B-Instruct-SFT-RL & 68.2& 90.4& 9.6& 46.0& 54.0& \xmark& \cmark \\
\rowcolor{color32}
Qwen2.5-14B-Instruct-SFT-RL & 74.6& 81.6& 18.4& 67.6& 32.4& \xmark& \cmark \\
\rowcolor{color32}
Qwen2.5-32B-Instruct-SFT-RL & 78.6& 88.0& 12.0& 69.2& 30.8& \xmark& \cmark \\
\bottomrule
\end{tabular}
\caption{\textbf{Performance on CriticLeanBench.}
The best, the second-best and the third-best scores for each indicator are shown in  \boxed{box}, \textbf{bold} and \underline{underlined}, respectively.}
\label{tab:main}
\end{table}
As indicated in Table \ref{tab:main}, the experimental results clearly demonstrate the effectiveness of the CriticLeanGPT models trained on our CriticLeanInstruct, in converting natural language mathematical statements into Lean 4 formal theorem declarations. Within the CriticLeanBench benchmark, our CriticLeanGPT models trained via supervised fine-tuning (SFT) and reinforcement learning (RL), along with their enhanced variants—outperform a range of closed-source API models, open-source models, and baseline models, highlighting distinct advantages. These outcomes yield several key insights: (1) Reasoning models excel in critical tasks, with Gemini 2.5 Pro, QwQ-32B, Qwen3-32B, and DeepSeek-R1 all attaining scores above 80. When compared to baseline models, our Qwen3-32B-RL model, optimized through RL, achieves a strong accuracy level, underscoring the efficacy of both our training methodology and dataset. 
(3) Our innovative mixed SFT strategy substantially boosts the performance of the Qwen2.5 family, with notable improvements observed across the 7B, 14B, and 32B models.
(4) Additionally, SFT and RL significantly strengthen the models' capacity to identify erroneous samples, as evidenced by higher true negative rates (TNR) and lower false negative rates (FNR)—a critical enhancement for accurate detection of incorrect formalizations, which is indispensable for effective critical tasks.

\subsection{Ablation Study}
\begin{table}[htbp]
\centering
\begin{tabular}{cccccc}
\toprule
Model & ACC & TPR & FPR & TNR & FNR \\
\midrule
\multicolumn{6}{c}{\textbf{\textit{7B Size Models}}} \\
\midrule
\rowcolor{color12}
Qwen2.5-7B-Instruct & 60.8& 89.6& 10.4& 32.0& 68.0 \\
\rowcolor{color12}
Qwen2.5-7B-Instruct-SFT(Critic Only) & 64.0& 70.8& 29.2& \boxed{57.2}& \boxed{42.8} \\
\rowcolor{color12}
Qwen2.5-7B-Instruct-SFT & \boxed{69.8}& \boxed{94.4}& \boxed{5.6}& 45.2& 54.8 \\
\midrule
\multicolumn{6}{c}{\textbf{\textit{14B Size Models}}} \\
\midrule
\rowcolor{color22}
Qwen2.5-14B-Instruct & 66.6& 80.4& 19.6& 52.8& 47.2 \\
\rowcolor{color22}
Qwen2.5-14B-Instruct-SFT(Critic Only) & 67.4& 80.8& 19.2& 54.0& 46.0 \\
\rowcolor{color22}
Qwen2.5-14B-Instruct-SFT & \boxed{70.6}& \boxed{83.6}& \boxed{16.4}& \boxed{57.6}& \boxed{42.4} \\
\midrule
\multicolumn{6}{c}{\textbf{\textit{32B Size Models}}} \\
\midrule
\rowcolor{color32}
Qwen2.5-32B-Instruct & 73.0& \boxed{91.6}& \boxed{8.4}& 54.4& 45.6 \\
\rowcolor{color32}
Qwen2.5-32B-Instruct-SFT(Critic Only) & 71.0& 72.0& 28.0& \boxed{70.0}& \boxed{30.0} \\
\rowcolor{color32}
Qwen2.5-32B-Instruct-SFT & \boxed{76.2}& 85.2& 14.8& 67.2& 32.8 \\
\bottomrule
\end{tabular}
\caption{\textbf{Comparison of model performance under different training strategies:} base model, SFT on Critic data only, and SFT on combined Critic, code, and math data. The best score for each indicator is shown in  \boxed{box}.}
\label{tab:mix}
\end{table}

\subsubsection{Effect of Reasoning Data}
We conduct an ablation study on CriticLeanBench to evaluate the impact of different training strategies. As shown in Table~\ref{tab:mix}, incorporating code and math reasoning data significantly improves performance across all model sizes compared to using Seed Data \ref{sec:seed_data}.
Specifically, using the CriticLeanInstruct \ref{sec:criticleaninstruct}, which samples Critic and non-Critic data at a ratio of 1:3, leads to substantial gains, demonstrating that integrating diverse reasoning tasks enhances the critical reasoning capabilities of the model. 
This suggests that multi-task learning with math and code data improves the critique abilities of mathematical formalization.

\subsubsection{Effect of SFT Dataset Size}
\begin{wrapfigure}{r}{0.5\textwidth}
    \centering
    \includegraphics[width=0.96\linewidth]{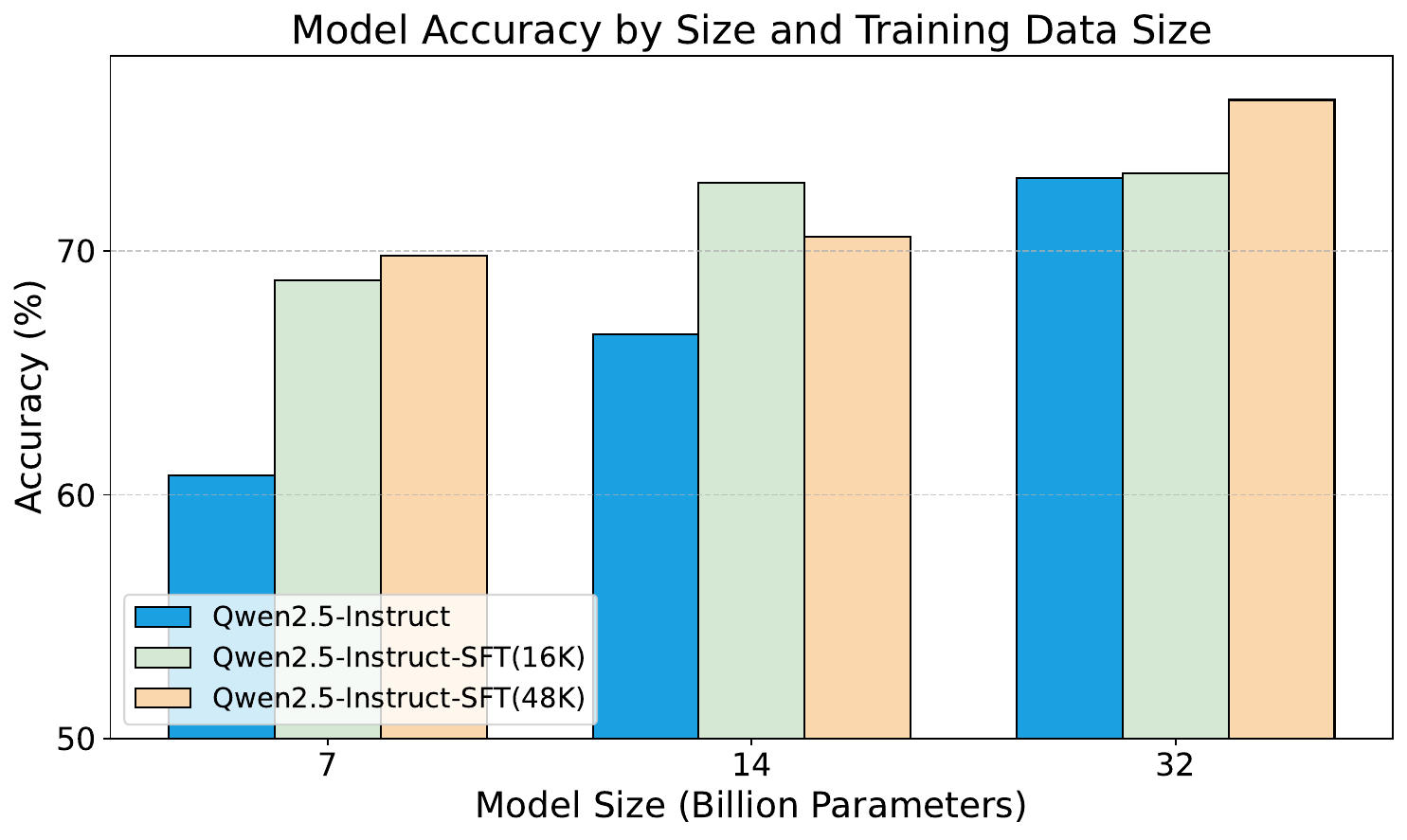}
    \caption{\textbf{Comparison of model performance under different amounts of SFT data:} base model, CriticLeanInstruct(16K), and CriticLeanInstruct.}
    \label{fig:sft_data}
    \vspace{-0.2in}  
\end{wrapfigure}

Figure \ref{fig:sft_data} highlights a notable parameter-dependent relationship between the size of the SFT dataset and key reasoning performance. Across all model scales, performance improvements are observed through SFT, albeit with varying degrees of fluctuation depending on the training set size. Smaller models (e.g., 7B) exhibit more pronounced gains as the dataset expands, whereas larger models (e.g., 32B) demonstrate a less consistent trend, with marginal improvements at lower data volumes but substantial gains at higher data volumes. These findings align with prior studies \citep{muennighoff2025s1, zhou2023lima}, underscoring the interplay between model capacity, data scale, and performance optimization in SFT scenarios. The results emphasize the need for tailored strategies to balance data efficiency and model generalization, particularly for large-scale architectures.

\subsection{Analysis}

\subsubsection{Scaling Analysis}

\begin{figure*}[h]
    \centering
    \includegraphics[width=1.0\linewidth]{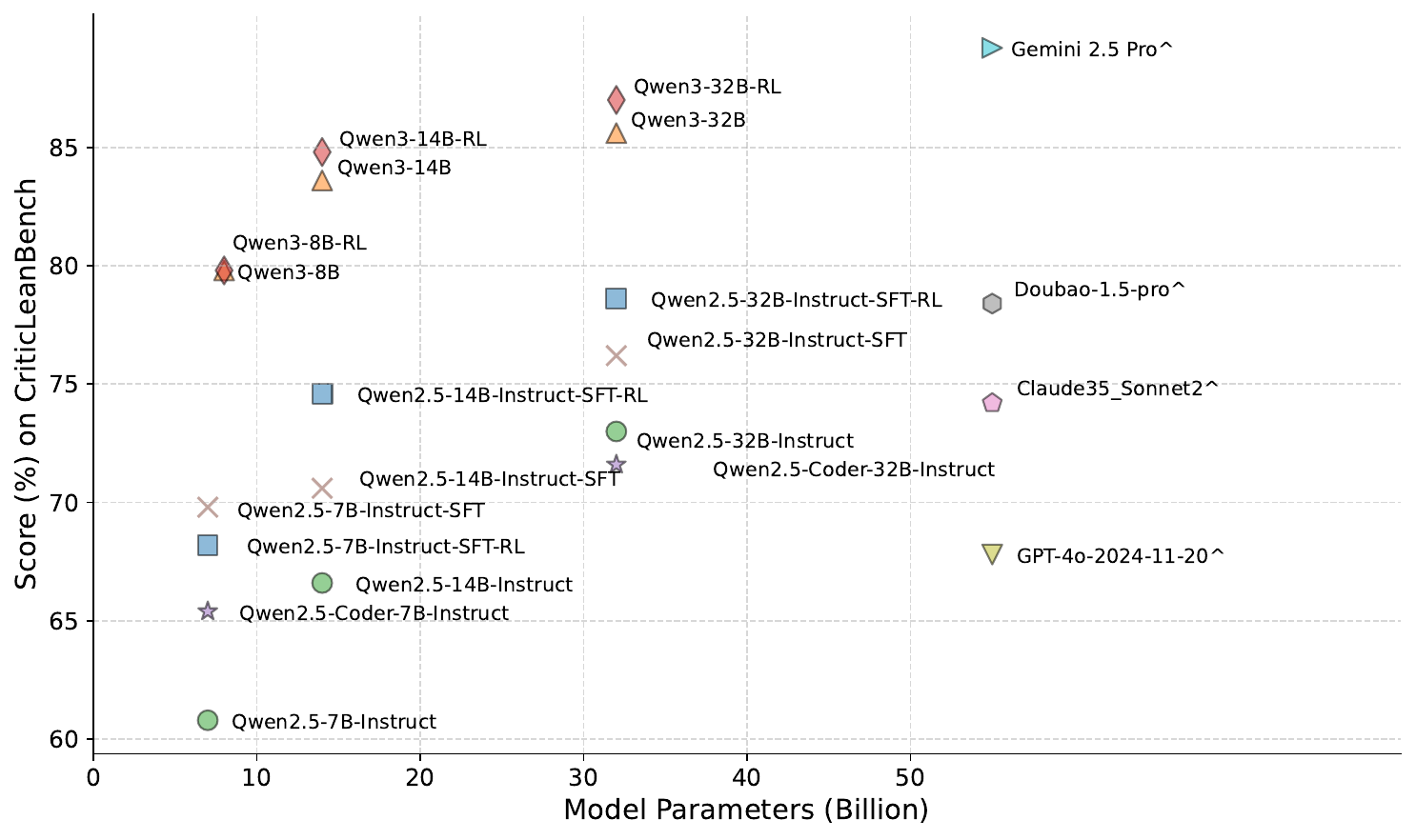}
    \caption{Scaling Analysis of LLMs on CriticLeanBench. \boxed{\textasciicircum{}} denoted closed-source LLMs.} \label{fig:scaling_law}
\end{figure*}

We evaluate the performance of Qwen series models on CriticLeanBench across different model scales, including Qwen2.5-Coder, Qwen2.5-Instruct, Qwen2.5-Instruct-SFT, Qwen2.5-Instruct-SFT-RL, Qwen3, and Qwen3-RL. The results in Figure \ref{fig:scaling_law} show that the performance improves consistently as the model size increases, demonstrating a clear scaling law of LLMs on CriticLeanBench.

\subsubsection{Effect of Pass@k}

\begin{figure*}[t]
    \centering
    \begin{subfigure}{0.48\linewidth}
        \includegraphics[width=\linewidth]{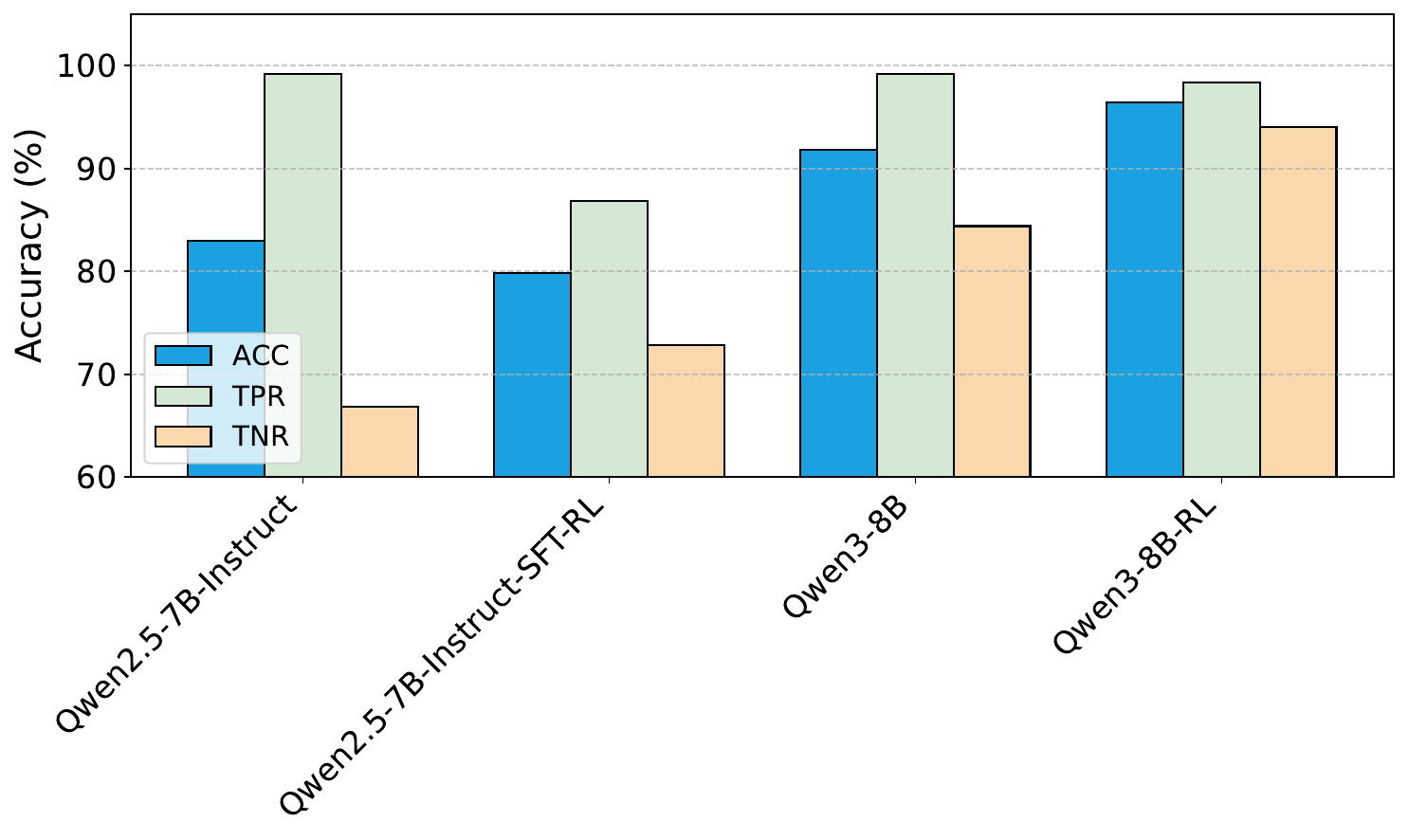}
        \label{fig:pass@8}
    \end{subfigure}
    \hfill
    \begin{subfigure}{0.48\linewidth}
        \includegraphics[width=\linewidth]{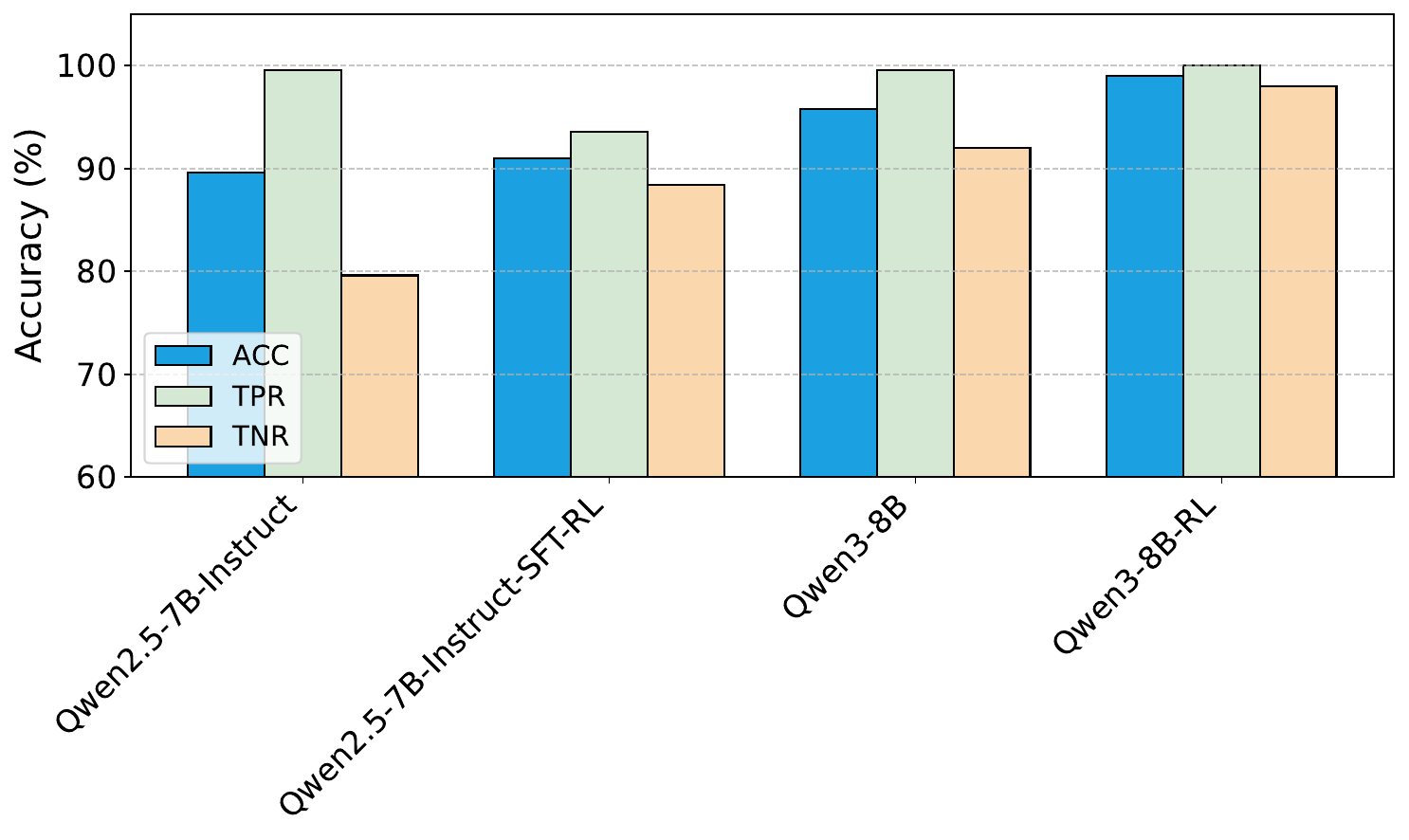}
        \label{fig:pass@32}
    \end{subfigure}

    \caption{Performance on CriticLeanBench using \textbf{Pass@k} metrics, where \textbf{k = 8} (left) and \textbf{k = 32} (right).}
    \label{fig:pass_all}
\end{figure*}

The Pass@k metric serves as a means to identify the highest-quality answers from multi-modal large language models (MLLMs) for final responses, highlighting the models' potential for enhancement through post-training techniques like Reinforcement Learning from Human Feedback (RLHF \citep{li2023reinforcement}) and Gradient Penalty Policy Optimization (GRPO \citep{shao2024deepseekmath}). In this study, Pass@k evaluations were conducted on Qwen2.5-7B-Instruct, Qwen2.5-7B-Instruct-SFT-RL, Qwen3-8B, and Qwen3-8B-RL using CriticLeanBench, with k set to 8 and 32, and evaluation metrics including Accuracy (ACC), True Positive Rate (TPR), and True Negative Rate (TNR).

As shown  in Figure \ref{fig:pass_all},
we observe that models subjected to Supervised Fine-Tuning (SFT) and Reinforcement Learning (RL) achieve better overall performance. Considering Pass@32, the Accuracy of Qwen2.5-7B-Instruct is outperformed by that of Qwen2.5-7B-Instruct-SFT-RL, which has undergone SFT and RL processing, leading to a notable improvement in the overall correctness rate. Regarding the True Negative Rate, models without SFT and RL processing are more prone to misclassifying incorrect samples as correct, whereas processed models effectively mitigate such errors; this is evident in the higher TNR of Qwen2.5-7B-Instruct-SFT-RL compared to Qwen2.5-7B-Instruct at Pass@32, reducing the likelihood of such misclassifications. A similar trend is observed in Qwen3-8B. Additionally, all models exhibit improved performance as k increases, suggesting that SFT- and RL-optimized models can more effectively select high-quality answers when provided with more candidate responses.


\begin{table}[t]
\centering
\begin{tabular}{lll}
\toprule 
\textbf{Dataset} & \textbf{Difficulty Level} & \textbf{Size} \\
\midrule
AOPs  ~\citep{aopsdataset}         & High School Olympiad & 199982 \\
DeepMath-103k~\citep{deepmath}            & Diverse  & 28538 \\
NuminaMath-TIR~\citep{numina_math_datasets}         & High School  & 22847 \\
DeepTheorem  ~\citep{zhang2025deeptheoremadvancingllmreasoning}         & High School Olympiad & 15131 \\
DeepScaleR~\citep{deepscaler2025}            & High School Olympiad & 11544 \\
DAPO-Math-17k~\citep{yu2025dapo}         & High School  & 4078 \\
Omni-MATH~\citep{gao2024omni} & Undergraduate        & 1176 \\
IneqMath~\citep{jiayi2025solving}         & High School Olympiad & 963 \\
BlueMO  ~\citep{bluemo2024} & High School Olympiad        & 616 \\
TAL-SCQ5K  ~\citep{TAL-SCQ5K}      & High School        & 393 \\
OnlineMathContest         & High School  & 70 \\
Multi-Source Math Competition  & High School Olympiad & 619 \\
\bottomrule
\end{tabular}
\caption{Overview of different sources for FineLeanCorpus.}
\label{tab:data-simplified}
\end{table}

\section{FineLeanCorpus}
\begin{wraptable}{r}{0.4\textwidth}  
\centering
\hspace{-5mm}  
\begin{tabular}{p{4cm} l}
\toprule
\textbf{Statistics} & \textbf{Number} \\
\midrule
\#Problems & 285957 \\
\addlinespace
\multicolumn{2}{l}{\textbf{Length}} \\
\hspace{1em}\textit{Statement} & \\
\hspace{2em}maximum length & 2980 tokens \\
\hspace{2em}minimum length & 9 tokens \\
\hspace{2em}avg length & 78.4 tokens \\
\hspace{1em}\textit{Lean Result(success)} & \\
\hspace{2em}maximum length & 768 tokens \\
\hspace{2em}minimum length & 14 tokens \\
\hspace{2em}avg length & 87.8 tokens \\

\bottomrule
\end{tabular}
\caption{Dataset statistics of FineLeanCorpus.}
\label{statics_fine_lean}
\end{wraptable}

To construct the FineLeanCorpus, we began by aggregating a vast and diverse collection of natural language mathematical problems. Sourcing from a wide array of materials, including high school olympiad datasets (e.g., AoPS, BlueMO), standard high school curricula (e.g., TAL-SCQ5), and undergraduate-level challenges (e.g., Omni-MATH), we ensured an extensive initial distribution in both the mathematical domain and difficulty. The first step in our process was to standardize this heterogeneous collection into a uniform, proof-based format, making each problem compatible with the Lean 4 theorem prover and ready for the subsequent formalization pipeline.

These standardized problems were then subjected to a rigorous, gated auto-formalization process powered by our CriticLean framework (Figure~\ref{fig:criticlean_intro}).The Kimina-Autoformalizer-7B model first generates a candidate formal statement. This statement must pass a syntactic check via the Lean 4 compiler; failure leads to regeneration. A successful compilation is followed by a semantic correctness check from our CriticLeanGPT model, with rejection also triggering a new attempt. This regenerative approach is critical, maximizing the yield from our source corpus by iteratively seeking a valid formalization. Finally, to further enhance precision, we apply a final filtering stage using another, higher-performance CriticLeanGPT model. Manual validation indicates that this step is expected to eliminate 74.7\% of the remaining incorrect formalizations. The resulting corpus is characterized by its expressive range: natural language statements vary from a concise 9 tokens to a complex 2,980 tokens (avg. 78.4), while their corresponding Lean formalizations span from 14 to 768 tokens (avg. 87.8), reflecting the deep spectrum of complexity successfully captured by our pipeline.

\begin{figure*}[h]
    \centering
    \includegraphics[width=1.0\linewidth]{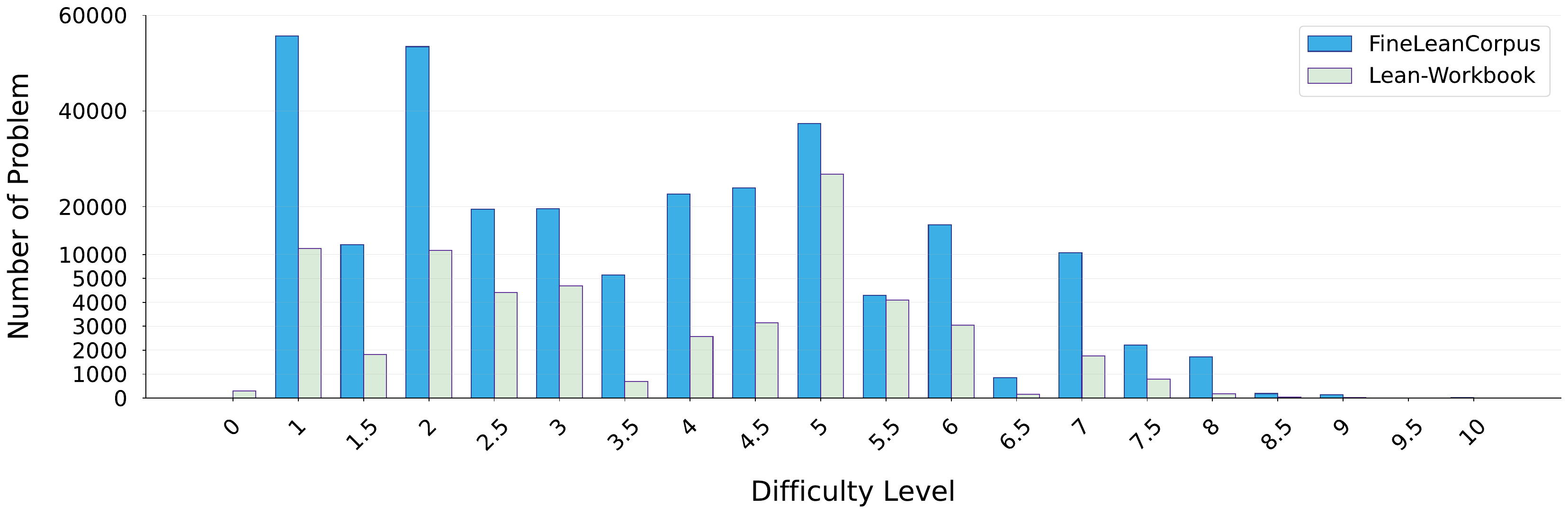}
    \caption{Comparison of dataset statistics. FineLeanCorpus offers a transparent critic process, a higher proportion of top-tier problems, and a more balanced and diverse topic distribution compared to the highly skewed Lean-Workbook.}\label{difficulty_distribution_compare}
\end{figure*}

\begin{figure*}[h]
    \centering
    \includegraphics[width=1.0\linewidth]{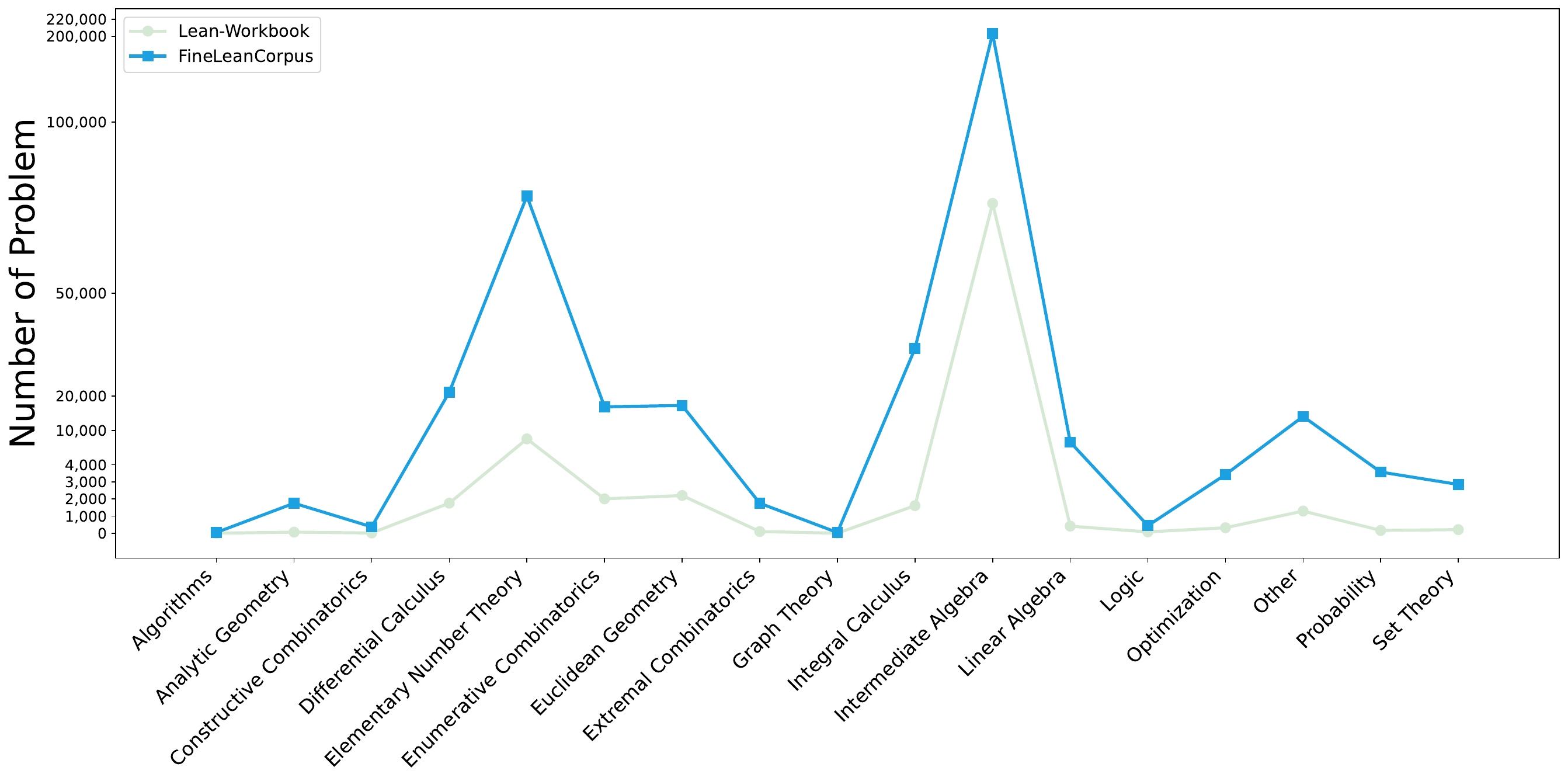}
    \caption{Math Domain Distributions: FineLeanCorpus  vs. Lean-Workbook.} \label{domain_second_layer_compare}
\end{figure*}

To further analyze the differences between our FineLeanCorpus dataset and the Lean-Workbook, we employed the templates from Appendix~\ref{appendix:Difficulty Level Assessment} and Appendix~\ref{appendix:Mathematical Problem Classification Task} to assess the difficulty levels and classify the mathematical domains of the datasets using Doubao-1.5-pro \citep{guo2025seed1}. A comparative analysis of our proposed FineLeanCorpus against Lean-Workbook (Figure~\ref{difficulty_distribution_compare}, Figure~\ref{domain_second_layer_compare}, Table~\ref{tab:data-simplified} and Table~\ref{statics_fine_lean}) reveals two fundamental advancements:
(1) Scale and Coverage: Our corpus provides a significant quantitative expansion in both problem difficulty and domain coverage. This expansion is evident not only across nearly the entire difficulty spectrum—offering a much richer data pool for training foundational skills—but also in the substantial augmentation of high-volume domains like Intermediate Algebra and Elementary Number Theory.
(2) FineLeanCorpus exhibits a more diverse and structurally balanced profile, achieved by collecting natural language statements from numerous sources. From a difficulty perspective, it features a multimodal distribution with substantial problem counts at several distinct complexity points, in stark contrast to the unimodal distribution of Lean-Workbook. This characteristic is crucial for mitigating model overfitting to a narrow complexity band. From a domain perspective, it substantially reinforces previously underrepresented areas. For instance, categories such as Analytic Geometry and Integral Calculus are significantly expanded, while niche topics like Algorithms and Graph Theory are also robustly augmented. This targeted enrichment transforms sparsely sampled topics into well-supported, learnable sub-domains, yielding a more comprehensive dataset designed to foster holistic reasoning capabilities. More details regarding our FineLeanCorpus dataset, including its fine-grained mathematical domain distribution (see Appendix~\ref{appendix:Fine-Grained Domain Distribution of FineLeanCorpus}), are provided for further analysis.

Furthermore, to push the boundaries of current models and foster research into the upper echelons of mathematical reasoning, we have curated a specialized training subset from FineLeanCorpus, which we designate as Diamond. This subset comprises 36,033 problems with a difficulty rating exceeding 5. The purpose of this high-difficulty training set is to create a demanding training environment that fosters the development of the sophisticated, multi-step reasoning required to tackle the most formidable mathematical problems.A detailed breakdown of the mathematical domain distribution within this subset is provided in table~\ref{tab:Fine-Grained Domain Distribution of FineLeanCorpus-Diamond}.

\begin{table}[htbp]
	\centering
	\resizebox{\textwidth}{!}{%
        \begin{tabular}{lccccccc} 
            \toprule
            \textbf{Dataset} & \textbf{Source} & \textbf{Theorems} & \textbf{Level} & \textbf{\begin{tabular}{c} Detailed \\ Critic Process \end{tabular}} & \textbf{\begin{tabular}{c} Difficulty \\ Profile \end{tabular}} & \textbf{\begin{tabular}{c} Topic \\ Diversity \end{tabular}} \\
            \midrule
            \textbf{Lean-Workbook}~\cite{ying2025leanworkbooklargescalelean} & Synthetic  & 140K & Undergraduate & Opaque & \begin{tabular}{c} Avg: 3.70 \\ Top-tier ($\ge$6): 7.81\% \end{tabular} & Highly Skewed \\
            
            \textbf{FineLeanCorpus (ours)} & Synthetic  & \textbf{285K} & Diverse & \textbf{Transparent} & \begin{tabular}{c} Avg: 3.24 \\ Top-tier ($\ge$6): \textbf{11.10\%} \end{tabular} & \textbf{Balanced \& Diverse} \\
            \bottomrule
        \end{tabular}%
    }
	\caption{Comparison of dataset statistics. FineLeanCorpus offers a transparent critic process, a higher proportion of top-tier problems, and a more balanced and diverse topic distribution compared to the highly skewed Lean-Workbook.}
	\label{tab:comparison_final_complete}
\end{table}




\subsection{Analysis on CriticLean Pipeline}
\begin{wraptable}{r}{0pt}  
\centering
\begin{tabular}{cc}
\toprule
Model & AVG \\
\midrule
\rowcolor{color12}
Kimina-Autoformalizer-7B & 38.0 \\
\rowcolor{color12}
Kimina-Autoformalizer-7B (Compiler) & 54.0 \\
\rowcolor{color42}
Kimina-Autoformalizer-7B (CriticLean) & \boxed{84.0} \\
\bottomrule
\end{tabular}
\caption{\textbf{Human evaluation accuracy results for autoformalization performance:} The best score is highlighted in \boxed{box}.}
\label{tab:autoformalization}
\end{wraptable}
As shown in Table~\ref{tab:autoformalization}, our autoformalization pipeline significantly improves accuracy. We selected 50 problems from the Omni-MATH and applied the following three formalization strategies, with the correctness of all outputs confirmed by manual human inspection.


Our baseline model, Kimina-Autoformalizer-7B, is used in each strategy:
(1) a \textbf{Single Pass} baseline (38.0\% accuracy);
(2) a \textbf{Compiler Feedback} loop, where the model regenerates formalizations until they successfully compile (54.0\% accuracy);
(3) our \textbf{CriticLean Pipeline}, which extends this process, regenerating formalizations until they both compile successfully and are validated by our integrated \textbf{CriticLeanGPT Model} (84.0\% accuracy).

\begin{table}[t]
  \centering
  \begin{tabular}{l cccccc} 
    \toprule
    \textbf{\# Attempt} & \textbf{1} & \textbf{5} & \textbf{10} & \textbf{50} & \textbf{100} & \textbf{200} \\
    \midrule
    Count / Ratio & 63 / 12.6\% & 137 / 27.4\% & 170 / 34.0\% & 229 / 45.8\% & 245 / 49.0\% & 264 / 52.8\% \\
    \bottomrule
  \end{tabular}
  \caption{ \textbf{Effectiveness of the Multi-Attempt Strategy on Formalization Yield}. 
           The table shows the cumulative number of successfully formalized problems retained by the critic model as the attempt limit increases. Statistics are from a 500-problem sample.}
             \label{tab:cumulative_yield}
\end{table}

\begin{table}[!ht]
\centering
\begin{tabular}{lclclc}
\toprule
\textbf{Dataset} & \textbf{Accuracy} & \textbf{Dataset} & \textbf{Accuracy}  & \textbf{Dataset} & \textbf{Accuracy}\\
\midrule
NuminaMath-TIR~\citep{numina_math_datasets}                    & 78\%                   & IneqMath~\citep{jiayi2025solving}             & 96\%                   &
BlueMO~\citep{bluemo2024}                 & 86\%           \\         DeepMath-103k~\citep{he2025deepmath}           & 84\%                   &
DAPO-Math-17k~\citep{yu2025dapo}              & 69\%                   & OnlineMathContest  & 88\%   \\                
AOPs~\citep{aopsdataset}                   & 73\%                   & TAL-SCQ5K~\citep{TAL-SCQ5K}          & 75\%                   &
Omni-MATH ~\citep{gao2024omni}   & 84\%                                    \\DeepTheorem~\citep{zhang2025deeptheoremadvancingllmreasoning}                   & 100\%                   & DeepScaleR~\citep{deepscaler2025}          & 100\%                   &
                          \\
\midrule
\end{tabular}
\caption{Human evaluation results of different sources.}
\label{tab:data_accuracy}
\end{table}

While compiler feedback resolves syntactical errors, it fails to detect logical flaws. Our pipeline addresses this gap. The integration of critic model, which performs deeper semantic and logical validation, is directly responsible for the accuracy increase from 54.0\% to 84.0\%. By filtering out plausible but incorrect formalizations, our method provides a more robust path toward reliable autoformalization. These manually-verified results present strong empirical evidence for the efficacy of our approach.


Table~\ref{tab:cumulative_yield} shows our pipeline achieved a 52.8\% success rate across 500 problems, where success required passing both syntactic validation and our critic model's semantic check. The value of our multi-attempt strategy is evident: while only 12.6\% of samples passed on the first try, the pipeline successfully recovered an additional 40.2\% that would be discarded by single-pass systems. Conversely, the 47.2\% failure rate within the 200-attempt limit highlights a fundamental bottleneck: the pipeline's performance is ultimately constrained by the base auto-formalization model's ability to produce a candidate our critic can approve.

Moreover, as shown in Table~\ref{tab:data_accuracy}, we also provide the human evaluation results of different sources. We observe that the accuracy varies across different sources, a discrepancy we attribute primarily to the differing domain and difficulty distributions of the respective datasets.
\footnote{Our human evaluation standard is particularly stringent. To calibrate our criteria, we inspected a random sample of 50 entries from the \texttt{Lean-Workbook}, which yielded an accuracy of 84\%.}

\section{Conclusion}
This paper presents CriticLean, a comprehensive framework that positions the critic as a central component in the autoformalization of mathematical statements. Through the development of CriticLeanGPT and the construction of CriticLeanBench, we demonstrate that explicitly modeling and training the critic yields significant improvements in formalization quality. Our pipeline not only refines the translation process through semantic validation, but also enables the construction of FineLeanCorpus,
which is validated by both compiler and critic. 

\newpage
\section{Contributions and Acknowledgments}

\subsection*{Co-First Authors}
Zhongyuan Peng, Yifan Yao, Kaijing Ma

\subsection*{Co-Authors}
Shuyue Guo, Yizhe Li, Yichi Zhang, Chenchen Zhang, Yifan Zhang, Zhouliang Yu, Luming Li, Minghao Liu, Yihang Xia, Jiawei Shen, Yuchen Wu, Yixin Cao, Zhaoxiang Zhang, Wenhao Huang

\subsection*{Corresponding Authors}
Jiaheng Liu, Ge Zhang

\clearpage

\bibliographystyle{plainnat}
\bibliography{main}

\clearpage

\beginappendix

\section{Fine-Grained Domain Distribution of FineLeanCorpus}
\label{appendix:Fine-Grained Domain Distribution of FineLeanCorpus}
The following table illustrates the fine-grained domain distribution of the FineLeanCorpus by presenting its 50 most frequent mathematical topics for illustrative purposes. To enhance readability and highlight the hierarchical structure, entries are sorted alphabetically by Main Category, then Sub-Category, and Topic.

\begin{longtable}{p{3cm} p{4cm} p{5cm} r}
\caption{Fine-Grained Domain Distribution of FineLeanCorpus} \label{tab:Fine-Grained Domain Distribution of FineLeanCorpus} \\
\toprule
\textbf{Main Category} & \textbf{Sub-Category} & \textbf{Topic} & \textbf{Count} \\
\midrule
\endfirsthead
\caption*{(Continued) Distribution of Problems by Mathematical Topic} \\
\toprule
\textbf{Main Category} & \textbf{Sub-Category} & \textbf{Topic} & \textbf{Count} \\
\midrule
\endhead
\bottomrule
\endfoot
\bottomrule
\endlastfoot

Algebra & Intermediate Algebra & Diophantine Equations & 23 \\
 &  & Functional Equations & 30178 \\
 &  & Inequalities & 75339 \\
 &  & Other & 65171 \\
 &  & Polynomials & 32424 \\
\cmidrule(l){2-4}
 & Linear Algebra & Matrices & 4361 \\
 &  & Other & 54 \\
 &  & Vector Spaces & 2179 \\
\midrule
Applied Mathematics & Algorithms & Greedy Algorithms & 46 \\
\cmidrule(l){2-4}
 & Optimization & Linear Programming & 258 \\
 &  & Other & 3165 \\
\cmidrule(l){2-4}
 & Other & Other & 273 \\
\cmidrule(l){2-4}
 & Probability & Conditional Probability & 363 \\
 &  & Expected Value & 792 \\
 &  & Other & 2417 \\
\midrule
Calculus & Differential Calculus & Applications of Derivatives & 57 \\
 &  & Derivatives & 14948 \\
 &  & Limits & 787 \\
 &  & Other & 5391 \\
\cmidrule(l){2-4}
 & Integral Calculus & Definite Integrals & 15165 \\
 &  & Other & 18761 \\
\cmidrule(l){2-4}
 & Other & Limits & 1086 \\
 &  & Limits of Multivariable Functions & 35 \\
 &  & Limits of Sequences & 106 \\
 &  & Other & 8592 \\
\midrule
Combinatorics & Constructive Combinatorics & Invariants & 320 \\
 &  & Other & 54 \\
\cmidrule(l){2-4}
 & Enumerative Combinatorics & Combinations & 5579 \\
 &  & Other & 10460 \\
 &  & Permutations & 869 \\
\cmidrule(l){2-4}
 & Extremal Combinatorics & Other & 375 \\
 &  & Pigeonhole Principle & 1377 \\
\cmidrule(l){2-4}
 & Graph Theory & Other & 42 \\
Discrete Mathematics & Logic & Propositional Logic & 444 \\
\cmidrule(l){2-4}
 & Other & Other & 3994 \\
\cmidrule(l){2-4}
 & Set Theory & Cardinality & 1262 \\
 &  & Other & 1585 \\
\midrule
Geometry & Analytic Geometry & Conic Sections & 1498 \\
 &  & Other & 254 \\
\cmidrule(l){2-4}
 & Euclidean Geometry & Circles & 1880 \\
 &  & Coordinate Geometry & 3584 \\
 &  & Other & 4172 \\
 &  & Transformations & 404 \\
 &  & Triangles & 7243 \\
\midrule
Number Theory & Elementary Number Theory & Diophantine Equations & 17208 \\
 &  & Divisibility & 21160 \\
 &  & Inequalities & 42 \\
 &  & Modular Arithmetic & 9284 \\
 &  & Other & 18636 \\
 &  & Prime Numbers & 12071 \\

\end{longtable}

\section{Fine-Grained Domain Distribution of FineLeanCorpus-Diamond}
The following table illustrates the fine-grained domain distribution of the Diamond dataset, our high-difficulty subset, by presenting its 50 most frequent mathematical topics for illustrative purposes. To enhance readability and highlight the hierarchical structure, entries are sorted alphabetically by Main Category, then Sub-Category, and Topic.
\begin{longtable}{p{3cm} p{4cm} p{5cm} r}
\caption{Fine-Grained Domain Distribution of FineLeanCorpus-Diamond} \label{tab:Fine-Grained Domain Distribution of FineLeanCorpus-Diamond} \\
\toprule
\textbf{Main Category} & \textbf{Sub-Category} & \textbf{Topic} & \textbf{Count} \\
\midrule
\endfirsthead
\caption*{(Continued) Distribution of Problems by Mathematical Topic} \\
\toprule
\textbf{Main Category} & \textbf{Sub-Category} & \textbf{Topic} & \textbf{Count} \\
\midrule
\endhead
\bottomrule
\endfoot
\bottomrule
\endlastfoot

Algebra & Intermediate Algebra & Functional Equations & 6732 \\
 &  & Inequalities & 12533 \\
 &  & Other & 3621 \\
 &  & Polynomials & 2798 \\
\cmidrule(l){2-4}
 & Linear Algebra & Matrices & 955 \\
 &  & Other & 7 \\
 &  & Vector Spaces & 206 \\
\midrule
Analysis & Real Analysis & Density of Sets & 3 \\
\midrule
Applied Mathematics & Optimization & Other & 108 \\
\cmidrule(l){2-4}
 & Other & Other & 9 \\
\cmidrule(l){2-4}
 & Probability & Conditional Probability & 9 \\
 &  & Expected Value & 23 \\
 &  & Other & 63 \\
Calculus & Differential Calculus & Derivatives & 913 \\
 &  & Differential Equations & 3 \\
 &  & Limits & 31 \\
 &  & Other & 375 \\
\cmidrule(l){2-4}
 & Integral Calculus & Definite Integrals & 2280 \\
 &  & Other & 2666 \\
\cmidrule(l){2-4}
 & Other & Limits & 44 \\
 &  & Other & 814 \\
\midrule
Combinatorics & Constructive Combinatorics & Invariants & 144 \\
 &  & Other & 33 \\
\cmidrule(l){2-4}
 & Enumerative Combinatorics & Combinations & 620 \\
 &  & Other & 1165 \\
 &  & Permutations & 70 \\
\cmidrule(l){2-4}
 & Extremal Combinatorics & Other & 201 \\
 &  & Pigeonhole Principle & 442 \\
\cmidrule(l){2-4}
 & Graph Theory & Other & 18 \\
 &  & Trees & 6 \\
\cmidrule(l){2-4}
 & Other & Other & 3 \\
\midrule
Discrete Mathematics & Logic & Propositional Logic & 21 \\
\cmidrule(l){2-4}
 & Other & Other & 1260 \\
\cmidrule(l){2-4}
 & Set Theory & Cardinality & 463 \\
 &  & Other & 379 \\
\midrule
Geometry & Analytic Geometry & Conic Sections & 29 \\
 &  & Other & 21 \\
\cmidrule(l){2-4}
 & Euclidean Geometry & Circles & 132 \\
 &  & Coordinate Geometry & 45 \\
 &  & Other & 306 \\
 &  & Transformations & 15 \\
 &  & Triangles & 1200 \\
\cmidrule(l){2-4}
 & Other & Other & 6 \\
\midrule
Number Theory & Elementary Number Theory & Diophantine Equations & 2106 \\
 &  & Divisibility & 3877 \\
 &  & Inequalities & 18 \\
 &  & Modular Arithmetic & 1593 \\
 &  & Other & 3523 \\
 &  & Prime Numbers & 3699 \\

\end{longtable}

\section{Formalization Quality Assessment Criteria}
\label{appendix:Formalization_Quality_Assessment_Criteria}
\begin{promptbox}[Formalization Quality Assessment Criteria]
\subsection*{I. Integrity \& Accuracy of Mathematical Content}
\begin{itemize}
    \item \textbf{Conditions \& Hypotheses:}  Are all explicit premises, variable domains (e.g., $\mathbb{N}$, $\mathbb{R}$, \texttt{Fin k}), index ranges (e.g., $a_0$ vs. $a_1$), properties of specific objects (e.g., geometric shapes, algebraic structures), and implicit context (e.g., non-zero divisors) accurately translated?  Are mathematical meanings preserved (e.g., $\neq 0$ vs. $> 0$)?
    \item \textbf{Goals \& Conclusions:}  Are all goals/conclusions translated (including multiple parts/cases)? Is the goal type accurate (e.g., specific value, extremum, existence/uniqueness)? For extrema, is attainability addressed?  Is the mathematical meaning accurately translated?
\end{itemize}

\subsection*{II. Clarity \& Correctness of Logical Structure}
\begin{itemize}
    \item \textbf{Propositional Structure:}  Are logical connectives ($\leftrightarrow$, $\rightarrow$, $\land$, $\lor$, $\neg$) and quantifiers ($\forall$, $\exists$, $\exists!$) used correctly?  Are quantifier order, scope, and nesting accurate (e.g., dependencies like $\forall \epsilon > 0, \exists \delta > 0, \dots$)?
    \item \textbf{Relation of Conditions to Conclusions:} How do multiple premises combine (e.g., $(A \land B) \rightarrow C$ vs. $A \rightarrow (B \rightarrow C)$)? Are constraints within the correct scope?
    \item \textbf{Reasoning Path:} Does the formalization reflect the original logic and key steps without introducing flaws or altering proof difficulty?
\end{itemize}

\subsection*{III. Lean Conventions \& Technical Accuracy}
\begin{itemize}
    \item \textbf{Syntax \& Declarations:} Is the syntax (e.g., parentheses, keywords, type declarations) correct? Are \texttt{theorem}, \texttt{example}, \texttt{lemma} used appropriately?
    \item \textbf{Type System:}  Do operations, parameters, and return values satisfy type constraints? Are numerals used in the correct type? Are mathematical concepts correctly mapped to Lean counterparts?
    \item \textbf{Definitions \& Library Usage:} Are custom definitions clear? Are imports correct and non-redundant? Are standard symbols and operations used correctly (e.g., \texttt{Complex.abs}, \texttt{Nat.Prime})?
    \item \textbf{Code Style \& Readability:}  Are names clear and consistent? Are there sufficient comments for complex parts?  Is there any redundancy?
\end{itemize}

\subsection*{IV. Problem Comprehension \& Overall Consistency}
\begin{itemize}
    \item \textbf{Grasping the Core:} Does the formalization capture the core mathematical idea?
    \item \textbf{Internal Self-Consistency:} Are there any logical contradictions between the translated parts?
    \item \textbf{Suitability for Formalization:} Is the problem suitable for formalization? Are assumptions/interpretations documented?
\end{itemize}

\subsection*{V. Formalization Strategy \& Choices}
\begin{itemize}
    \item \textbf{Abstraction Level:} Is the abstraction level appropriate, avoiding unnecessary generalization or over-specification?
    \item \textbf{Alternative Evaluation:} Were alternatives considered?  Was decomposition/modularization used for complex problems?
\end{itemize}

\subsection*{VI. Provability \& Proof Assistance}
\begin{itemize}
    \item \textbf{Proof Complexity:} Does the formalization maintain a similar proof complexity? Were lemmas added to simplify the proof?
    \item \textbf{Automation Potential:} Is the structure amenable to automation tools?
\end{itemize}
\end{promptbox}
\section{Error Taxonomy} 
\label{appendix:Error Taxonomy}
\begin{promptbox}[Error Taxonomy ]

\textbf{1. Semantic and Logical Errors}

\begin{enumerate}
\item[\textbf{1.1}] \textbf{Premise Translation Error}
\begin{itemize}
\item \textbf{Description:} This error occurs when formalizing the given conditions, constraints, or assumptions from the original problem, resulting in a discrepancy between the logical premises in the Lean code and the problem's description.
\item \textbf{Examples:} Failing to constrain variables to be positive, integers, or coprime as required; not ensuring that denominators in mathematical expressions are non-zero; omitting geometric constraints such as "A, B, C form a triangle" in geometry problems; not explicitly specifying the exact range of a variable. For instance, relationships between angles and sides must be precisely defined in Lean, otherwise the resulting correspondence may not be unique.
\end{itemize}

Generated code
\item[\textbf{1.2}] \textbf{Mathematical Representation Error}
\begin{itemize}
    \item \textbf{Description:} This error involves an inaccurate representation of the form and meaning of mathematical entities such as variables and expressions from the original problem. This leads to the formalized mathematical proposition in the code being inconsistent with the original problem, thereby undermining formal semantic correspondence.
    \item \textbf{Examples:} Formalizing a cubic polynomial as a quadratic one; mistranslating "all eigenvalues are 1" as "the determinant is 1"; simplifying a complex algebraic relationship into an incorrect equation; using the conclusion as a premise; mismatch of mathematical entities. An incorrect expression structure can change the intrinsic structure and semantics of the original mathematical expression, even if the computed result might happen to be the same. For example, incorrectly formalizing a finite nested expression (e.g., $2002+21(2001+\dots)$) as the sum of an infinite series.
\end{itemize}

\item[\textbf{1.3}] \textbf{Goal Translation Error}
\begin{itemize}
    \item \textbf{Description:} The final goal or conclusion achieved by the code does not match what the problem asks for, failing to complete the specified task.
    \item \textbf{Examples:} The problem asks for a specific numerical value, but the code only proves its existence; the problem asks to calculate the radius of convergence, but the code incorrectly solves for the sum of the series; the final answer has a numerical calculation error or a formal writing error (e.g., writing a fraction $n/m$ as $m/n$). The final goal might also be translated incompletely, with omissions.
\end{itemize}

\item[\textbf{1.4}] \textbf{Variable Usage Error}
\begin{itemize}
    \item \textbf{Description:} Improper use of a variable's type, scope, name, or index.
    \item \textbf{Examples:} Using natural numbers ($\mathbb{N}$) for a variable that requires real numbers ($\mathbb{R}$); off-by-one errors in summation or sequence indices; confusing or redefining variable names.
\end{itemize}

\item[\textbf{1.5}] \textbf{Misuse of Mathematical Concepts}
\begin{itemize}
    \item \textbf{Description:} Incorrectly using a mathematical formalism in Lean to represent a different mathematical concept.
    \item \textbf{Examples:} Translating "calculate the residue" as "find the limit"; formalizing "locally uniform convergence" as "pointwise convergence"; treating a problem of counting unordered combinations (e.g., non-congruent triangles) as one of counting ordered tuples.
\end{itemize}

\item[\textbf{1.6}] \textbf{Incorrect Assumption}
\begin{itemize}
    \item \textbf{Description:} Adding conditions that are not present in the original problem, which oversimplifies the problem or leads to an incorrect conclusion.
    \item \textbf{Example:} Introducing an unfounded assumption, such as a specific numerical value.
\end{itemize}

\end{enumerate}

\textbf{2. Lean Syntax and Technical Errors}
\begin{itemize}
\item \textbf{Description:} These are technical issues at the code level that prevent the code from compiling or cause unexpected runtime behavior.
\end{itemize}
\begin{enumerate}
\item[\textbf{2.1}] \textbf{Syntax Error}
\begin{itemize}
\item \textbf{Description:} The code does not conform to the basic syntax rules of Lean 4.
\item \textbf{Examples:} A \texttt{theorem} statement is missing its name; the \texttt{by sorry} block to skip a proof is absent; incorrect keywords or symbols are used.
\end{itemize}

Generated code
\item[\textbf{2.2}] \textbf{Type Error}
\begin{itemize}
    \item \textbf{Description:} Performing incompatible operations on variables of different data types, including type mismatches and type casting errors.
    \item \textbf{Examples:} Performing division on a natural number (\texttt{Nat}) and expecting a fractional result, but the outcome is floored to 0; failing to cast integers or natural numbers to real numbers before performing real-valued operations.
\end{itemize}

\item[\textbf{2.3}] \textbf{Operator \& Parenthesis Error}
\begin{itemize}
    \item \textbf{Description:} The calculation order of an expression does not match the intended logic due to misunderstandings of operator precedence, improper placement of quantifiers, or incorrect use of parentheses.
    \item \textbf{Example:} $\tan^2(\frac{\pi}{9})$ is incorrectly parsed as $(\frac{\tan\pi}{9})^2$.
\end{itemize}

\item[\textbf{2.4}] \textbf{Library Usage Error}
\begin{itemize}
    \item \textbf{Description:} Improper use of functions or definitions from \texttt{mathlib}.
    \item \textbf{Examples:} Using \texttt{.ncard} on an incorrect type of set; using a deprecated function name like \texttt{Complex.abs}.
\end{itemize}

\end{enumerate}

\textbf{3. Translation Completeness and Other Meta-Errors}
\begin{itemize}
\item \textbf{Description:} These errors reflect that the formalization fails to cover all requirements of the problem, or that the problem itself is difficult to formalize.
\end{itemize}
\begin{enumerate}
\item[\textbf{3.1}] \textbf{Unformalizable Problem}
\begin{itemize}
\item \textbf{Description:} The original problem description is vague, ambiguous, relies on diagrams, or involves real-world scenarios that are difficult to express in formal logic.
\item \textbf{Examples:} The problem depends on a geometric figure that is not explicitly defined; a physical context or narrative scenario cannot be modeled precisely; the problem statement itself contains mathematical errors.
\end{itemize}

Generated code
\item[\textbf{3.2}] \textbf{Incomplete Formalization}
\begin{itemize}
    \item \textbf{Description:} The code only formalizes part of the problem, omitting other requirements.
    \item \textbf{Examples:} Ignoring a "prove or disprove" requirement and assuming the statement is true by default; omitting multi-step derivation requirements from the problem.
\end{itemize}

\item[\textbf{3.3}] \textbf{Improvable Code Style}
\begin{itemize}
    \item \textbf{Description:} The code may be logically correct but can be improved in terms of clarity, robustness, or adherence to conventions.
    \item \textbf{Examples:} Adding parentheses could enhance logical clarity; variable names are not intuitive; better use could be made of Lean's syntactic features.
\end{itemize}

\end{enumerate}

\end{promptbox}
\section{Complete Checklist for Lean4 Mathematical Formalization} \label{appendix:checklist}

\begin{promptbox}[Complete Checklist for Lean4 Mathematical Formalization]
\label{appendix:Complete Checklist for Lean4 Mathematical Formalization}
\textbf{Conditions \& Hypotheses:}
\begin{enumerate}
\item{Completeness of Preconditions: Are all explicitly stated preconditions in the problem translated without omission?}
\item{Accuracy of Variable Domains: Are the domains of variables (e.g., $\mathbb{N}, \mathbb{N}^+, \mathbb{R}$, \leanterm{Fin k}, \leanterm{Set.Icc a b}) accurately translated?}
\item{Accuracy of Indexing: Do the starting points and ranges of sequence/function indices (e.g., $a_0$ vs $a_1$, \leanterm{Finset.range n} vs \leanterm{Finset.Icc 1 n}) align with the original intent?}
\item{Clarity of Object Properties: Are the properties of specific objects (e.g., geometric figures like trapezoids, incircles; algebraic structures like groups, rings) clearly expressed?}
\item{Inclusion of Implicit Conditions: Are common implicit contextual conditions in mathematics (e.g., non-zero divisors, non-negative radicands, non-degenerate geometric objects, definedness of functions/sequences at application points, default to real numbers if unspecified) appropriately added?}
\item{Accuracy of Conditional Semantics: Is the mathematical meaning of conditions (e.g., "not equal to 0" ($\neq 0$) vs "greater than 0" ($> 0$), direction of inequality signs ($>$ vs $\ge$)) accurately translated?}
\end{enumerate}

\textbf{Goals \& Conclusions:}
\begin{enumerate}
\item{Completeness of Goals/Conclusions: Are all goals/conclusions that need to be proven or solved translated? (Pay special attention to multi-part conclusions and multiple solution scenarios).}
\item{Precision of Goal Type: Is the type of goal to be solved precise (e.g., specific value, extremum, existence/uniqueness, universal property, equivalence relation, implication)?}
\item{Attainability in Extremum Problems: For extremum problems, is "attainability" explicitly stated (i.e., demonstrating not just an inequality, but also that equality can be achieved)?}
\item{Accuracy of Goal Semantics: Is the mathematical meaning of the goals accurately translated?}
\end{enumerate}

Combination of Preconditions

\textbf{Logical Structure:}
\begin{enumerate}
\item{Accuracy of Logical Connectives: Does the use of logical connectives ($\leftrightarrow$ (iff), $\rightarrow$ (if...then...), $\land$ (and), $\lor$ (or), $\neg$ (not)) accurately reflect the logical relationships of the original proposition?}
\item{Appropriateness of Quantifiers: Is the use of quantifiers ($\forall$ (for all), $\exists$ (exists), $\exists!$ (exists uniquely)) appropriate?}
\item{Correctness of Quantifier Scope and Nesting: Do the order, scope, and nesting of quantifiers correctly express the dependencies between variables (e.g., in $\forall \epsilon > 0, \exists \delta > 0, \ldots, \delta$ depends on $\epsilon$)?}
\item{Combination of Preconditions: How do multiple preconditions combine to affect the conclusion (e.g., differentiate $(A \land B) \rightarrow C$ from $A \rightarrow (B \rightarrow C)$)?}
\item{Fidelity to Original Logic: Does the formalization faithfully represent the inherent logic and key steps of the original mathematical problem?}
\end{enumerate}

\textbf{Lean Technical Accuracy:}
\begin{enumerate}
\item{Correctness of Basic Syntax: Is the basic Lean syntax (parenthesis matching, keywords like \leanterm{theorem}, \leanterm{def}, \leanterm{variable}, \leanterm{let}, \leanterm{by}) entirely correct?}
\item{Adherence to Type Constraints: Do all operations, function parameters, and return values satisfy Lean's type constraints?}
\item{Correct Mapping of Mathematical Concepts: Are mathematical concepts correctly mapped to their Lean counterparts?}
\item{Clarity of Custom Definitions: Are all custom functions, predicates, and notations used clearly defined?}
\item{Correctness of Imports: Are necessary definitions and lemmas correctly imported from Mathlib?}
\end{enumerate}

\textbf{Overall Consistency:}
\begin{enumerate}
\item{Capturing Core Mathematical Ideas: Does the formalization truly capture the core mathematical ideas and goals of the original problem?}
\item{Absence of Logical Contradictions: Are there any logical contradictions between the translated conditions, definitions, and goals?}
\item{Appropriateness for Formalization: Is the problem itself suitable for precise, unambiguous mathematical formalization?}
\item{Documentation of Assumptions: Are any assumptions or interpretative choices made during the formalization process documented?}
\end{enumerate}

\end{promptbox}

\section{Prompt:Critical Feedback to CoT}
\label{appendix:Critical_Feedback_to_CoT}
\begin{promptbox}[Prompt:Critical Feedback to CoT]

\textbf{Instruction:}You will be provided with a mathematical text and its Lean4 code representation. Your task is to evaluate whether the Lean4 code accurately and semantically represents the mathematical text. You will be given a boolean indicating conversion success and potentially failure information. Based on this, use a step-by-step Chain of Thought (COT) to generate a detailed explanation for why the conversion is considered successful or failed, focusing on the semantic equivalence and formal correctness of the Lean4 code relative to the mathematical meaning. The Lean4 code must preserve the intended meaning in the mathematical text and use correct Lean4 syntax and structure.

\textbf{Input: You will receive the following four values:}
\begin{enumerate}
    \item \textbf{Mathematical Text}: A string containing mathematical content.  
    \item \textbf{Lean4Code}: A string representing the code equivalent of the mathematical text.
    \item \textbf{Conversion Success}: A boolean value (True or False) indicating whether the mathematical text was successfully converted to the Lean4 code representation.
    \item \textbf{Reason}: A string representing the code equivalent of the mathematical text. 
        \item If Conversion Success is True, this field will typically be empty. You must generate the detailed justification for the success.
        \item If Conversion Success is False, this field will contain specific information pinpointing why the conversion failed. You must elaborate on this identified failure, incorporating the analysis modules described below.
\end{enumerate}

\textbf{Your Role:}
You are an AI language assistant. Your role is to analyze the provided information and, using a step-by-step Chain of Thought (COT) approach, generate the explanation for the conversion's success (if indicated as successful) or elaborate on the identified failure information (if indicated as failed).
\textbf{Guidelines:}
\begin{enumerate}
\item \textbf{Understand the Content:} Carefully read the mathematical text, the code representation, the Conversion Success value, and the Reason input (if Conversion Success is False).

\item \textbf{Generating the Explanation for Success (if Conversion Success is True):}
        If the conversion is successful, provide a detailed, step-by-step explanation using COT to justify why it is successful. Your justification should implicitly cover:
        \begin{enumerate} 
            \item \textbf{Mathematical Text Analysis:} Briefly identify the core mathematical components (definitions, variables, operations, relations, constraints, statements) present in the text.
            \item \textbf{Lean4 Code Analysis:} Briefly describe the structure and components of the provided Lean4 code, outlining how it represents the mathematical elements.
            \item \textbf{Comparative Analysis:} Systematically compare each mathematical component identified in the text analysis with its corresponding part described in the Lean4 code analysis. Explain step-by-step how the Lean4 syntax and constructs accurately translate the mathematical concept and semantics.
            \item \textbf{Confirmation of Correctness:} Explicitly confirm the absence of errors by verifying that:
            \begin{itemize} 
                \item Definitions: Types (variables, functions, sets, etc.) are correctly defined in Lean4, match the mathematical context, accurately reflect the intended meaning, and have a valid interpretation within Lean4. (Absence of Errors in Definition)
                \item Constraints: All constraints from the mathematical text are present (no omissions), accurately represented (no inconsistencies or logical flaws), and without unnecessary additions (no redundancy). (Absence of Constraint Errors)
                \item Syntax: The Lean4 code uses correct syntax for logical expressions (quantifiers, connectives), mathematical expressions ($\sum$, $\int$, \{\}, operations), and overall Lean4 language structure (theorem declarations, keywords like by). (Absence of Syntax Errors)
                \item Proof Targets/Statements: If applicable, the proof target or statement in Lean4 is complete, unambiguous, and consistent with the mathematical claim. (Absence of Proof Target Errors)
                \item Confirm that the Lean4 code fully and accurately captures the mathematical meaning without loss or distortion, preserving the intended meaning. Highlight the key aspects demonstrating a correct and complete translation.
            \end{itemize}
        \end{enumerate}

\item \textbf{Elaborating on Failure (if Conversion Success is False):}
        If the conversion is unsuccessful, use a step-by-step COT approach structured around the following analysis modules to elaborate on the specific failure identified in the input Reason field:
        \begin{enumerate} 
            \item \textbf{Mathematical Text Analysis:} Briefly identify the core mathematical components (definitions, variables, operations, relations, constraints, statements) present in the text. This establishes the intended meaning.
            \item \textbf{Lean4 Code Analysis:} Briefly describe the structure and components of the provided Lean4 code, outlining how it *attempts* to represent the mathematical elements identified in the text analysis step. This describes the code *as presented*, before focusing on the error.
            \item \textbf{Comparative Analysis:} Compare the intended meaning and structure derived from the mathematical text analysis with the actual Lean4 code structure described in the previous step. \textbf{Crucially, use the input \texttt{Reason} to pinpoint and focus the comparison on the specific segment(s) where the Lean4 code fails to accurately represent the mathematical text}, clearly demonstrating the mismatch or divergence indicated by the failure reason.
            \item \textbf{Identification of Omission/Error:} Based on the discrepancy identified in the comparative analysis (which was guided by the input \texttt{Reason}), clearly articulate the specific error. Categorize this error by referencing the relevant error type and explain why the issue constitutes this type of error. Use the following categories and examples as a guide:
            \begin{itemize} 
                \item Errors in Definition: e.g., Incorrect or mismatched type definitions (Type Mismatch), failure to reflect the precise mathematical context (Context Mismatch), definitions being mathematically meaningless or ill-formed in Lean4 (Ill-formed Definition).
                \item Constraint Errors: e.g., Omission of necessary constraints (Constraint Omission), Redundancy of constraints (Constraint Redundancy), Inconsistency with mathematical text (Constraint Inconsistency), Logical Errors within the translated constraints (Logical Flaw in Constraint).
                \item Syntax Errors: e.g., Errors in Logical Expression Syntax (quantifiers $\forall$, $\exists$; connectives $\land$, $\lor$, $\to$, $\neg$), Mathematical Expression Syntax (notation for sums $\sum$, integrals $\int$, sets \{\}, specific operations), or Lean4 Language Structure Syntax (keywords like \texttt{theorem}, \texttt{def}, \texttt{variable}, \texttt{assume}, \texttt{show}, \texttt{by}; overall declaration structure). (Invalid Lean Syntax, Logical Syntax Error, Mathematical Notation Error)
                \item Proof Target Errors: e.g., Complete Omission of the target statement (Target Omission), Partial Omission or ambiguity in the target (Target Incomplete/Ambiguous), Inconsistency between the Lean4 goal and the mathematical claim (Target Mismatch).
            \end{itemize}
            \item \textbf{Explanation of Impact:} Detail the consequences of this specific error. Explain how it alters the logical meaning compared to the original mathematical text, introduces ambiguity, makes the statement fundamentally different, or renders the Lean4 code syntactically invalid or semantically incorrect relative to the intended mathematical statement.
        \end{enumerate}

\item \textbf{Crucially:} Your final explanation should present this analysis directly. Do not explicitly state "The provided reason indicates..." or similar phrases referring back to the input Reason field in your output. Simply explain the error based on the structure above.
        \begin{itemize} 
        \item  \textbf{Focus on Explanation:} Your primary goal is to generate a clear and comprehensive explanation using the COT method, ensuring each step in your reasoning is explicit.
        \item  \textbf{No Evaluation of Failure Information:} If the conversion failed, do not question or evaluate the validity of the failure information given in the Reason input. Your task is solely to elaborate on it based on that information, following the specified structure and error categorization.
        \end{itemize}
\end{enumerate} 

\textbf{Output Format:}
The output must be in English and presented strictly as follows:
\begin{verbatim}
Reason: [Your step-by-step Chain of Thought explanation here, either 
justifying success or elaborating on the identified failure following 
the specified modules and error types for failures]
\end{verbatim}
This section will contain a single paragraph with your detailed Chain of Thought explanation.

\textbf{Notes:}
\begin{itemize}
    \item Clarity and Conciseness: Ensure your explanation is clear, logical, and easy to understand. While using COT, ensure each step is articulated concisely.
    \item Professional Tone: Maintain an objective and professional tone throughout your response.
    \item No Additional Information: Do not introduce any external information or perform the conversion yourself. Focus solely on generating the explanation based on the provided inputs, using a step-by-step approach guided by the structure and error categories provided.
\end{itemize}

\end{promptbox}

\section{Prompt:Formal Verification Expert}
\label{appendix:Formal Verification}
\begin{promptbox}[Prompt:Formal Verification Expert]
Role: Lean \& Formal Verification Expert \\
\\
Input:

- Mathematical\_Text: A math problem and its answer (no proof).\\
- Lean4Code: A Lean 4 theorem statement formalizing the problem. Proof is intentionally omitted (e.g., sorry).\\
\\
Goal:\\
Determine if the Lean theorem statement is an exact and faithful formalization of the mathematical problem.\\  
**Do not evaluate or consider the answer or the proof. Your sole task is to verify the correctness of the formalization.**\\
\\
Evaluation Stages (All required):\\
\\
1. Math Assertion Analysis \\ 
   Identify all structurally and semantically relevant components of the mathematical problem, including variables, types, quantifiers, constraints, logic structure, conclusion, and so on. The analysis should be based on the actual content of the text.\\
\\
2. Lean Statement Analysis (ignore proof part)\\  
   Extract all structurally and semantically relevant components from the Lean statement, including variables, types, conditions, quantifiers, constraints, the final claim, and so on. The analysis should reflect the actual content present in the Lean code.\\
\\
3. Comparative Verification  \\
   Check for exact correspondence between the math and Lean statements; you may refer to aspects like:\\
   - Semantic alignment, logic structure, and quantifier correctness.\\
   - Preservation of constraints and boundary assumptions.\\
   - Accurate typing and use of variables.\\
   - Syntactic validity and proper Lean usage (free from errors).\\
   - Use of symbols and constructs without semantic drift.\\
   - No missing elements, no unjustified additions, and no automatic corrections or completions.\\
\\
4. Final Judgement  
   Based solely on the above analysis, judge whether the Lean statement is a correct and exact formalization of the mathematical problem.\\
\\
5. Accuracy Confirmation  \\
   If correct: clearly confirm why all elements match.  \\
   If incorrect: list all mismatches and explain how each one affects correctness.\\
\\
Note: While the analysis may be broad and open to interpreting all relevant features, the final judgment must be based only on what is explicitly and formally expressed in the Lean statement.  \\
**Do not consider or assess any part of the proof. Your judgment should be entirely about the accuracy of the statement formalization.**\\
\\
Output Format:\\
Return exactly one JSON object:

\textbf{```json}
\begin{verbatim}
{
  "reasons": "Your detailed CoT analysis:
1. Math Assertion Analysis: [...]
2. Lean Statement Analysis (Proof Ignored): [...]
3. Comparative Verification: [...]
4. Conclusion: [...]
5. Accuracy Confirmation: [match confirmation or list of discrepancies...]",
  "is_assistant_correct": "[Correct/Incorrect]"
}
\end{verbatim}
\textbf{```}

\medskip 
Input Data:\\
— Start of Mathematical\_Text —\\
\{\text{mathematical\_statement}\}\\
— End of Mathematical\_Text —\\
\\
— Start of Lean4Code —\\
\{\text{autoformalization\_placeholder}\}\\
— End of Lean4Code —

\end{promptbox}
\section{Prompt:Lean Flaw Injection}
\label{appendix:Lean Flaw Injection}

\begin{promptbox}[Prompt:Lean Flaw Injection]
You are an exceptional Lean 4 code formalization engineer. Your current task is to meticulously introduce errors into correct Lean 4 code, following a specific checklist of error types.

I will provide you with:
\begin{enumerate}
    \item {A problem pair consisting of:}
    \begin{enumerate}
        \item A mathematical definition or statement.
        \item Its corresponding correct Lean 4 code formalization.
    \end{enumerate}
    \item{An error checklist and Lean 4 theorem statement:} A list of potential error types or modification strategies, along with a Lean 4 theorem statement that formalizes the problem. Note that the proof is intentionally omitted (e.g., using \texttt{sorry}).
\end{enumerate}

Your process should be as follows:
\begin{enumerate}
    \item  From the provided checklist, select exactly 2 error types or modification strategies. Each chosen item must be directly and plausibly applicable to the structure, logic, or types within the provided mathematical statement and its Lean 4 code.
    \item  Based on your selection(s), intentionally modify the  Lean 4 code to make it incorrect or subtly deviate from the original mathematical intent. The modification should be contextually relevant to the provided mathematical statement and its original Lean 4 formalization. Aim for errors that someone might plausibly make when formalizing *this specific* concept, rather than completely arbitrary changes. 
    \item  When applying multiple selected error types, aim to incorporate all of them naturally into the code modification. If this proves too complex or makes the resulting error contrived, you may focus on a primary subset of the selected types, but clearly explain your rationale and how the chosen modifications relate to your selections.
    \item Important: Do not add any comments directly within the mathematical description or the Lean 4 code itself to explain your changes. All explanations should be in the "Detailed Explanation of Modifications" section.
\end{enumerate}
\subsection*{}
You must then return your response as a JSON object with the following structure and content:

\textbf{```json}




\textbf{Output Format}

You must then return your response as a JSON object with the following structure and content:
\begin{verbatim}
{
  "modified_lean_code": "[Your intentionally flawed Lean 4 code here. 
  Please ensure that you do not add any comments directly within the 
  mathematical description or the Lean 4 code itself to explain your 
  changes.]",
  "explanation": "Here, you will:\n1. Clearly state which 2 error 
  types or modification strategies you selected from the checklist.
  \n2. Explain precisely what changes you made to the original 
  correct code to introduce these errors/deviations.\n3. Describe 
  how these changes reflect the selected strateg(ies). Crucially,
  point out the specific \"error points\" you introduced by comparing
  the modified code to the (unstated) original correct version, and 
  explain *why* this type of error is plausible or relevant given 
  the specific mathematical content and structure of the original 
  code."
}
\end{verbatim}

\medskip 
\textbf{checklist:}

\{\text{selected\_checklist\_items\_placeholder}\}\par 
"Mathematical statement": "\{\text{refined\_statement\_placeholder}\}"\par
"Lean4 code": "\{\text{autoformalization\_placeholder}\}"\par

\end{promptbox}
\section{Prompt: Mathematical Problem Classification Task}
\label{appendix:Mathematical Problem Classification Task}

\begin{promptbox}[Prompt: Mathematical Problem Classification Task]
I am working with natural language statements of advanced mathematics problems, which are intended to be later formalized in Lean.
Before formalization, I need to categorize each problem into its appropriate mathematical domain.

\textbf{\#OBJECTIVE\#}
\begin{enumerate}
    \item Summarize the math problem in one or two sentences, highlighting the key mathematical concepts or structures involved.
    \item Classify the problem into one or more mathematical domains, using a hierarchical classification chain. For example: \\
    \texttt{Algebra -> Intermediate Algebra -> Inequalities}.
\end{enumerate}

The classification should be based on the mathematical content of the \textit{natural language} problem, even if no formal notation is used yet.

\textbf{\#DOMAIN TAXONOMY\#}

The domain classification should follow a standard taxonomy:

\texttt{<math domains>}\\
Algebra -> Intermediate Algebra -> Inequalities\\
Algebra -> Intermediate Algebra -> Polynomials\\
Algebra -> Intermediate Algebra -> Functional Equations\\
Algebra -> Linear Algebra -> Vector Spaces\\
Algebra -> Linear Algebra -> Matrices\\
Geometry -> Euclidean Geometry -> Triangles\\
Geometry -> Euclidean Geometry -> Circles\\
Geometry -> Euclidean Geometry -> Coordinate Geometry\\
Geometry -> Euclidean Geometry -> Transformations\\
Geometry -> Analytic Geometry -> Conic Sections\\
Number Theory -> Elementary Number Theory -> Divisibility\\
Number Theory -> Elementary Number Theory -> Modular Arithmetic\\
Number Theory -> Elementary Number Theory -> Diophantine Equations\\
Number Theory -> Elementary Number Theory -> Prime Numbers\\
Combinatorics -> Enumerative Combinatorics -> Permutations\\
Combinatorics -> Enumerative Combinatorics -> Combinations\\
Combinatorics -> Extremal Combinatorics -> Pigeonhole Principle\\
Combinatorics -> Constructive Combinatorics -> Invariants\\
Combinatorics -> Graph Theory -> Trees\\
Calculus -> Differential Calculus -> Derivatives\\
Calculus -> Integral Calculus -> Definite Integrals\\
Discrete Mathematics -> Logic -> Propositional Logic\\
Discrete Mathematics -> Set Theory -> Cardinality\\
Applied Mathematics -> Probability -> Expected Value\\
Applied Mathematics -> Probability -> Conditional Probability\\
Applied Mathematics -> Optimization -> Linear Programming\\
Applied Mathematics -> Algorithms -> Greedy Algorithms\\
Algebra -> Intermediate Algebra -> Other\\
Geometry -> Euclidean Geometry -> Other\\
Number Theory -> Elementary Number Theory -> Other\\
Combinatorics -> Enumerative Combinatorics -> Other\\
Calculus -> Integral Calculus -> Other\\
Discrete Mathematics -> Other -> Other\\
\texttt{</math domains>}

\textbf{\#RESPONSE FORMAT\#}
\textbf{\#\#Summarization}
[A brief summary of the problem, describing the mathematical concepts it involves.]

\textbf{\#\#Math domains}
[A hierarchical classification of the mathematical domains involved in the problem.]

\textbf{\#INSTRUCTIONS\#}
\begin{itemize}
    \item You may include up to \textbf{three} domain classification chains, separated by semicolons.
    \item The format must be: \texttt{Major Domain -> Subdomain -> Specific Topic}.
    \item If a concept doesn't fit exactly, use \texttt{Other} as the last node only. For example: \texttt{Algebra -> Intermediate Algebra -> Other}.
    \item Avoid using vague or overlapping categories.
    \item End each report with the line \texttt{=== report over ===}.
\end{itemize}

\textbf{\#INPUT\#}
Below is the original natural language math problem statement:

\texttt{<statement>\{statement\}</statement>}

\textbf{\#OUTPUT FORMAT\#}
Please return your response in \textbf{JSON} format. For example:
\begin{verbatim}
{
  "Summary": "This problem involves minimizing a symmetric function 
  over real variables under absolute value constraints.",
  "Domains": [
    "Algebra -> Intermediate Algebra -> Inequalities",
    "Calculus -> Differential Calculus -> Applications of Derivatives"
  ]
}
\end{verbatim}
\end{promptbox}

\section{Prompt:Difficulty Level Assessment}
\label{appendix:Difficulty Level Assessment}

\begin{promptbox}[Prompt:Difficulty Level Assessment]
You are an exceptional Lean 4 code formalization engineer. Your current task is to meticulously introduce errors into correct Lean 4 code, following a specific checklist of error types.

I am working with natural language statements of advanced mathematics problems, which are intended to be later formalized in Lean.
Before that, we aim to assess the **intrinsic difficulty** of the problem in its current informal (natural language) form.

\textbf{\#OBJECTIVE}

Assign a **difficulty score** to the problem, on a scale from 0 to 10.

Your rating should reflect the mathematical reasoning required to solve the problem — including the level of abstraction, creativity, number of steps, and familiarity with advanced techniques.

\textbf{\#DIFFICULTY REFERENCE}

\textbf{\#\#Examples for difficulty levels}

For reference, here are problems from each of the difficulty levels 1-10:

\noindent\textbf{1:} How many integer values of x satisfy $|x| < 3\pi$? (2021 Spring AMC 10B, Problem 1)

\noindent\textbf{1.5:} A number is called flippy if its digits alternate between two distinct digits. For example, 2020 and 37373 are flippy, but 3883 and 123123 are not. How many five-digit flippy numbers are divisible by 15? (2020 AMC 8, Problem 19)

\noindent\textbf{2:} A fair 6-sided die is repeatedly rolled until an odd number appears. What is the probability that every even number appears at least once before the first occurrence of an odd number? (2021 Spring AMC 10B, Problem 18)

\noindent\textbf{2.5:} A, B, C are three piles of rocks. The mean weight of the rocks in A is 40 pounds, the mean weight of the rocks in B is 50 pounds, the mean weight of the rocks in the combined piles A and B is 43 pounds, and the mean weight of the rocks in the combined piles A and C is 44 pounds. What is the greatest possible integer value for the mean in pounds of the rocks in the combined piles B and C? (2013 AMC 12A, Problem 16)

\noindent\textbf{3:} Triangle $\triangle ABC$ with $AB = 50$ and $AC = 10$ has area 120. Let D be the midpoint of AB, and let E be the midpoint of AC. The angle bisector of $\angle BAC$ intersects DE and BC at F and G, respectively. What is the area of quadrilateral FDBG? (2018 AMC 10A, Problem 24)

\noindent\textbf{3.5:} Find the number of integer values of $k$ in the closed interval $[-500, 500]$ for which the equation $\log(kx) = 2 \log(x + 2)$ has exactly one real solution. (2017 AIME II, Problem 7)

\noindent\textbf{4:} Define a sequence recursively by $x_0 = 5$ and
\[ x_{n+1} = \frac{x_n^2 + 5x_n + 4}{x_n + 6} \]
for all nonnegative integers $n$. Let $m$ be the least positive integer such that
\[ x_m \le 4 + \frac{1}{2^{20}}. \]
In which of the following intervals does $m$ lie? (A) [9, 26] (B) [27, 80] (C) [81, 242] (D) [243, 728] (E) [729, $\infty$) (2019 AMC 10B, Problem 24 and 2019 AMC 12B, Problem 22)

\noindent\textbf{4.5:} Find, with proof, all positive integers $n$ for which $2^n + 12^n + 2011^n$ is a perfect square. (USAJMO 2011/1)

\noindent\textbf{5:} Find all triples $(a, b, c)$ of real numbers such that the following system holds:
\[ a + b + c = \frac{1}{a} + \frac{1}{b} + \frac{1}{c} , \]
\[ a^2 + b^2 + c^2 = \frac{1}{a^2} + \frac{1}{b^2} + \frac{1}{c^2}. \]
(JBMO 2020/1)

\noindent\textbf{5.5:} Triangle ABC has $\angle BAC = 60^\circ$, $\angle CBA \le 90^\circ$, $BC = 1$, and $AC \ge AB$. Let H, I, and O be the orthocenter, incenter, and circumcenter of $\triangle ABC$, respectively. Assume that the area of pentagon BCOIH is the maximum possible. What is $\angle CBA$? (2011 AMC 12A, Problem 25)

\noindent\textbf{6:} Let $\triangle ABC$ be an acute triangle with circumcircle $\omega$, and let H be the intersection of the altitudes of $\triangle ABC$. Suppose the tangent to the circumcircle of $\triangle HBC$ at H intersects $\omega$ at points X and Y with $HA = 3$, $HX = 2$, and $HY = 6$. The area of $\triangle ABC$ can be written in the form $m\sqrt{n}$, where $m$ and $n$ are positive integers, and $n$ is not divisible by the square of any prime. Find $m + n$. (2020 AIME I, Problem 15)

\noindent\textbf{6.5:} Rectangles $BCC_1B_2$, $CAA_1C_2$, and $ABB_1A_2$ are erected outside an acute triangle ABC. Suppose that $\angle BC_1C + \angle CA_1A + \angle AB_1B = 180^\circ$.
Prove that lines $B_1C_2$, $C_1A_2$, and $A_1B_2$ are concurrent. (USAMO 2021/1, USAJMO 2021/2)

\noindent\textbf{7:} We say that a finite set $S$ in the plane is balanced if, for any two different points $A, B$ in $S$, there is a point $C$ in $S$ such that $AC = BC$. We say that $S$ is centre-free if for any three points $A, B, C$ in $S$, there is no point $P$ in $S$ such that $PA = PB = PC$. Show that for all integers $n \ge 3$, there exists a balanced set consisting of $n$ points. Determine all integers $n \ge 3$ for which there exists a balanced centre-free set consisting of $n$ points. (IMO 2015/1)

\noindent\textbf{7.5:} Let $\mathbb{Z}$ be the set of integers. Find all functions $f : \mathbb{Z} \to \mathbb{Z}$ such that
\[ x f(2f(y) - x) + y^2 f(2x - f(y)) = f(x)^2 + f(y f(y)) \]
for all $x, y \in \mathbb{Z}$ with $x \ne 0$. (USAMO 2014/2)

\noindent\textbf{8:} For each positive integer $n$, the Bank of Cape Town issues coins of denomination $1/n$. Given a finite collection of such coins (of not necessarily different denominations) with total value at most $99 + 1/2$, prove that it is possible to split this collection into 100 or fewer groups, such that each group has total value at most 1. (IMO 2014/5)

\noindent\textbf{8.5:} Let I be the incentre of acute triangle ABC with $AB \ne AC$. The incircle $\omega$ of ABC is tangent to sides BC, CA, and AB at D, E, and F, respectively. The line through D perpendicular to EF meets $\omega$ at R. Line AR meets $\omega$ again at P. The circumcircles of triangle PCE and PBF meet again at Q. Prove that lines DI and PQ meet on the line through A perpendicular to AI. (IMO 2019/6)

\noindent\textbf{9:} Let $k$ be a positive integer and let $S$ be a finite set of odd prime numbers. Prove that there is at most one way (up to rotation and reflection) to place the elements of $S$ around the circle such that the product of any two neighbors is of the form $x^2 + x + k$ for some positive integer $x$. (IMO 2022/3)

\noindent\textbf{9.5:} An anti-Pascal triangle is an equilateral triangular array of numbers such that, except for the numbers in the bottom row, each number is the absolute value of the difference of the two numbers immediately below it. For example, the following is an anti-Pascal triangle with four rows which contains every integer from 1 to 10.
\begin{center}
4 \\
2 \quad 6 \\
5 \quad 7 \quad 1 \\
8 \quad 3 \quad 10 \quad 9
\end{center}
Does there exist an anti-Pascal triangle with 2018 rows which contains every integer from 1 to $1 + 2 + 3 + \dots + 2018$? (IMO 2018/3)

\noindent\textbf{10:} Prove that there exists a positive constant $c$ such that the following statement is true: Consider an integer $n > 1$, and a set $S$ of $n$ points in the plane such that the distance between any two different points in $S$ is at least 1. It follows that there is a line $\ell$ separating $S$ such that the distance from any point of $S$ to $\ell$ is at least $cn^{-1/3}$.

\textbf{\#OBJECTIVE}
\begin{enumerate}
    \item Summarize the math problem in a brief sentence, describing the concepts involved in the math problem.
    \item Based on the source of the given problem, as well as the difficulty of the problems referenced in these materials and the solution to the current problem, please provide an overall difficulty score for the current problem. The score should be a number between 1 and 10, with increments of 0.5, and should align perfectly with the materials.
\end{enumerate}

\textbf{\#STYLE\#}

Data report.

\textbf{\#TONE\#}

Professional, scientific.

\textbf{\#AUDIENCE\#}

Students. Enable them to better understand the difficulty of the math problems.

\textbf{\#RESPONSE: MARKDOWN REPORT\#}
\textbf{\#\#Summarization}

[Summarize the math problem in a brief paragraph.]

\textbf{\#\#Difficulty}

[Rate the difficulty of the math problem and give the reason.]

\textbf{\#ATTENTION\#}
\begin{itemize}
    \item Add "=== report over ===" at the end of the report.
\end{itemize}

\textbf{\#INPUT\#}
Below is the original natural language math problem statement:

<statement>\{statement\}</statement>

\textbf{\#OUTPUT FORMAT\#}

You must respond with a JSON object:

\begin{verbatim}
{
  "Difficulty": float (between 0 and 10),
  "Rationale": "Explain your score in 1–3 sentences. Mention structural 
  elements or comparison to benchmark problems."
}
\end{verbatim}

\end{promptbox}

\end{document}